\UseRawInputEncoding 
\documentclass[graybox]{svmult}
\usepackage{subfiles}
\usepackage{blindtext}

\usepackage{type1cm}      
\usepackage{makeidx}        
\usepackage{graphicx}        
\usepackage{multicol}         
\usepackage[bottom]{footmisc} 
\usepackage{newtxtext}      
\usepackage{newtxmath}     
\usepackage{changes}
\makeindex            

\usepackage{booktabs}
\usepackage{color}
\definecolor{Blue}{rgb}{0,0,1}

\definecolor{Green}{rgb}{0,1,0}

\begin{document}

\title*{Multi-label Ranking: Mining Multi-label and Label Ranking Data}
 
\author{Lihi Dery}
 
\institute{Lihi Dery \at Ariel University, Ariel. Israel, \email{lihid@ariel.ac.il}}
 
\maketitle

\abstract{We survey multi-label ranking tasks, specifically multi-label classification and label ranking classification. We highlight the unique challenges, and re-categorize the methods, as they no longer fit into the traditional categories of transformation and adaptation. We survey developments in the last demi-decade, with a special focus on state-of-the-art methods in deep learning multi-label mining, extreme multi-label classification and label ranking. We conclude by offering a few future research directions.}

%{Each chapter should be preceded by an abstract (no more than 200 words) that summarizes the content. The abstract will appear \textit{online} at \url{www.SpringerLink.com} and be available with unrestricted access. This allows unregistered users to read the abstract as a teaser for the complete chapter.
%Please use the 'starred' version of the \texttt{abstract} command for typesetting the text of the online abstracts (cf. source file of this chapter template \texttt{abstract}) and include them with the source files of your manuscript. Use the plain \texttt{abstract} command if the abstract is also to appear in the printed version of the book.}

\section{Introduction}\label{sec:intro}

Multi-label ranking (MLR) is the problem of predicting and ranking multiple labels for a single instance. The predicted labels are known as the instance's \textit{labelset}. 
MLR can be typically reduced to two sub-problems: 
%changed 28.4.2020
The first is multi-label classification, where the task is to bipartite the data into relevant labels (the labelset) and irrelevant labels. 
The second is label ranking classification, where the task is to rank labels for each instance. A label ranking may contain ties; in the extreme case relevant labels hold a tie on first place, and irrelevant labels hold a tie on second place, thus turning the label ranking classification into a multi-label one. 
%original
%the first is to rank labels for each instance, and the second is to place a threshold on the ranked list in order to bipartite the data into relevant and irrelevant labels. 
%The ranking may contain ties, in the extreme case relevant labels have a tie on first place, and irrelevant labels have a tie on second place. The labels predicted for each instance are known as the instance's \textit{labelset}. 

The first studies of MLR originated from text categorization problems, where each document was labeled with several predefined topics \cite{joachims1998text, mccallum1999multi, schapire2000boostexter}.
The labels were sometimes ranked in order of importance \cite{ioannou2010}.
While multi-label for text categorization continues to be an active field \cite{mironczuk2018, mujtaba2019}, 
multi-label ranking has spread to many more domains. In bioinformatics, a gene can belong to multiple functional families \cite{clare2001}. In music, a tune can spark many emotions \cite{trohidis2008multi}. In medical diagnosis, an x-ray image can have multiple labels \cite{baltruschat2019comparison}. 
In social networks, people may belong to several interest groups \cite{wang2013multi} and
in visual object recognition, objects can be ordered according to their relevance to the picture \cite{bucak2009efficient, yang2016exploit}.
%In audio clips ? (see zhang's 2013 review maybe?)
%In crowd-sourcing, multiple interpretations of the same example may be given.

There are six common challenges in MLR that either do not exist in single-label classification or are intensified in MLR settings.

\begin{figure}[h!]
\centering
\includegraphics[width=0.7\textwidth]{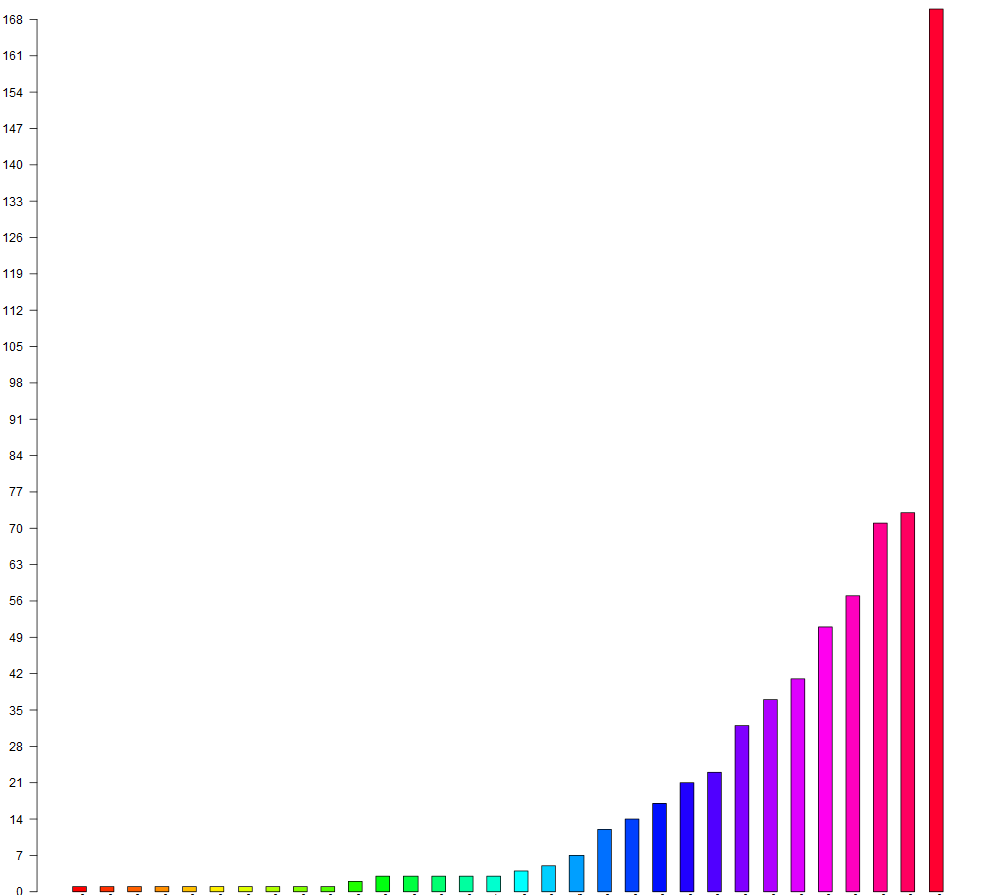}
\caption{High dimensionality in the output space. The number of instances (y-axis) with a given labelset (x-axis) in the Genebase dataset.}
\label{figure:fig1}
\end{figure}

\begin{figure}[h!]
\centering
\includegraphics[width=0.75\textwidth]{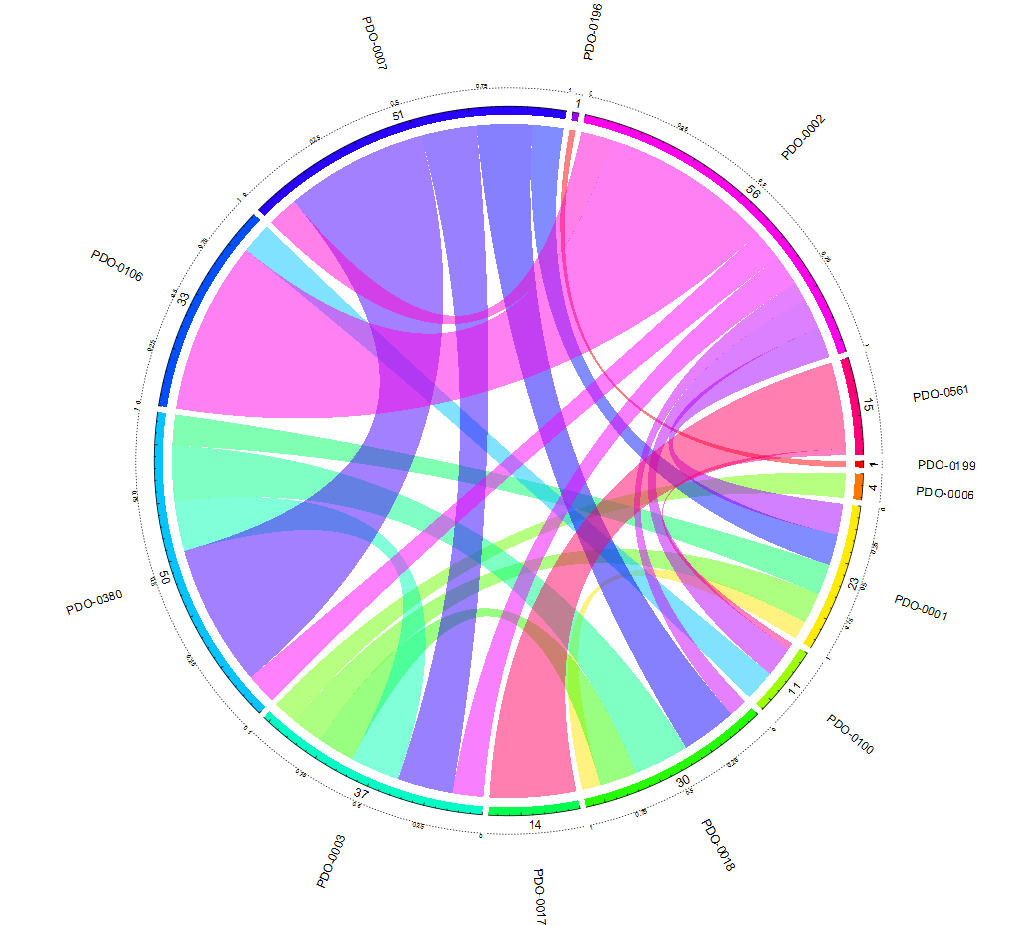}
\caption{Label correlation: a chord diagram \cite{gu2014circlize} showing the concurrence of 13 labels in the Genebase dataset. The arcs represent label concurrence For example, the protein to the right of 12 o'clock (PDO-0196) co-occurs with only one other protein (PDO-0199) and that happens less frequently than other concurrences.}
\label{figure:fig2}
\end{figure}

\begin{figure}[h!]
\centering
\includegraphics[width=0.8\textwidth]{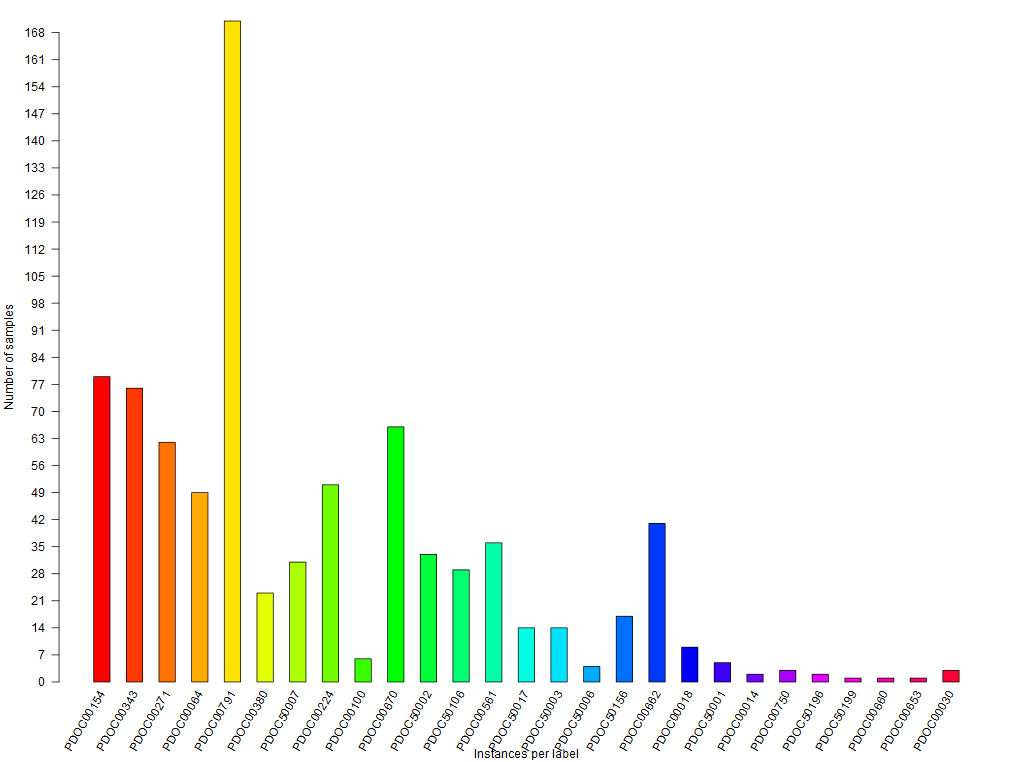}
\caption{Label imbalance. The number of instances (y-axis) with a given label (x-axis) in the Genebase dataset.}
\label{figure:fig3}
\end{figure}

\begin{figure}[h!]
\centering
\includegraphics[width=0.75\textwidth]{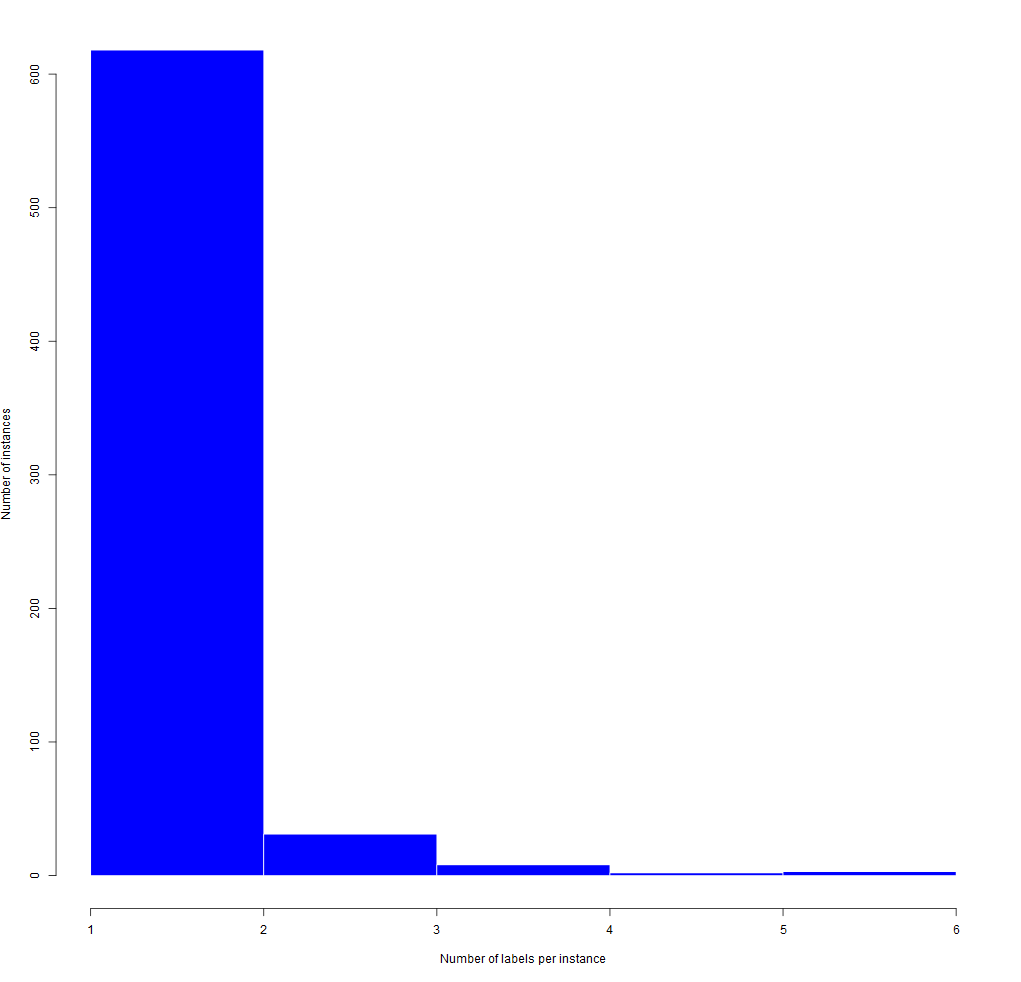}
\caption{Labelset size imbalance. The number of instances (y-axis) with a given number of labels (x-axis) in the Genebase dataset.}
\label{figure:fig4}
\end{figure}

\begin{itemize}
    \item \textbf{High dimensionality in the output space.} High dimensionality in the \textit{input space} (i.e., data with millions of instances) or in the \textit{feature space }(i.e., data with thousands or millions of features) is also common in single-label classification, though in MLR it might be harder to solve. However, high dimensionality in the\textit{ output space} is unique to MLR. The number of possible labelsets (i.e. label combinations) grows exponentially with the number of labels. This often leads to sparseness of available data and to class imbalance, as some labelsets may appear often, while others may be rare or may not appear in the training set at all. See an example in figure \ref{figure:fig1}.
    \item \textbf{Label correlation.} This aspect is fundamental in MLR. If there are no relations between the labels, the problem can be split into multiple binary classification problems without loss of information. However, in MLR, the relation between the labels is complex. For example, if two labels have a high concurrence, the model is supposed to somehow boost the prediction of one label, if the other is predicted. If two labels have a parallel relation, i.e., they do not concur, the model is expected to handle that as well. See an example of label concurrence in figure \ref{figure:fig2}. 
    \item \textbf{Label imbalance.} The label distribution is highly skewed, most labels have only a handful of positive training instances and a few labels dominate with many training instances. See an example in figure \ref{figure:fig3}. %(few-shot)
    \item \textbf{Labelset size imbalance.} The labelset size of each instance is highly skewed \cite{rubin2012statistical}. Few instances have much more labels than average, while most instances have very few labels. See an example in figure \ref{figure:fig4}. %Power-law (long tail)
    \item \textbf{Label importance.} Not all labels are equally important to the characterization of the instance. The label importance is explicitly known if the target class input labelset is ranked (i.e, if a ranked labels are provided as input in the training data). Otherwise, the label importance remains to be inferred. 
    \item \textbf{Zero-shot labels and labelsets.} Some labels and labelsets never appear in the training set. 
\end{itemize}

Figures \ref{figure:fig1} - \ref{figure:fig4}, were created using the mldrGUI \cite{charte-charte:2015} with the Genebase dataset \cite{diplaris2005protein} as an example. The dataset contains the classification of proteins into families with similar function. Each instance is a protein, the attributes are the protein's motifs. The labels are the families the protein belongs to. This is a multi-label setting since each protein can belong to more than one family. In it's current online version \cite{mulan}, the dataset contains 662 instances (the proteins) with 1213 attributes each. There are 27 possible labels (families) and 32 possible labelsets.

%Co-occurrences of labels along with frequency of occurrences are visualized in XXX. 

%These problems may also appear in regular MLR problems, but in XML problems it is more acute.
%An example few-shot and long-tail  distributions is portrayed by Papanikolaou et al. \cite{papanikolaou2017large} for the MEDLINE database, the premier bibliographic database of the National Library of Medicine. The articles in this database are labelled with concepts of the MeSH (Medical Subject Headings) ontology (also curated by NLM).

Previous literature reviews on multi-label methods \cite{gibaja2014multi, gibaja2015tutorial, herrera2016multilabel, zhang2013review}  and label ranking methods \cite{vembu2010label, zhou2014taxonomy}, while excellent, are slightly outdated.  
%and do not cover the most recent developments such as extreme MLR and deep learning multi-label algorithms. 
Our contributions in this survey are three-fold.
First, we refresh the definition of MLR \cite{brinker2006unified} and define it's two sub-tasks: multi-label classification and label ranking classification. 
Second, we suggest to re-categorize MLR methods, as they no longer fit 
into the traditional categories of transformation or adaptation. We thus suggest new categories such as deep learning multi-label methods, extreme multi-label methods and label ranking methods.
Third, we focus on the last demi-decade which has not yet been surveyed. 

The rest of this survey is organized as follows:
we first define MLR and place it in context with other tasks (section \ref{sec:definition}).  
Next,we survey recent developments in multi-label methods, with a special focus on deep learning methods and the emerging field of extreme multi-label classification (section \ref{sec:ml}). We then move on to discuss label ranking (section \ref{sec:lr}). We present up to date information about evaluation (section \ref{sec:eval}) and conclude by offering a few research directions on open problems (section \ref{sec:future}). 
 
%We offer a few insights: we classify the algorithms according to evaluation metric, we classify the problems tasks (MLR, LR, ML etc) according to their features.

\section{Definition and context}\label{sec:definition}

We begin with a definition of MLR. Next we place MLR in context by detailing which problems can be seen as sub cases of MLR, and what MLR is a sub problem of. 

\begin{definition}\label{def:mlr}
\textbf{Multi-label ranking (MLR).} An MLR task is characterized by $\textbf{x} \in \mathcal{X}$ 
instances and $l \in L$ labels with the following properties:
\begin{enumerate}
    \item $\mathcal{X}$ is finite and contains $n$ instances.
    \item $L$ is finite and contains $m$ labels.
    \item An ordered labelset $Y = [\,l_1,...,l_q]\,$ with $q \leq |L|$ labels, 
     contains a subset of the $L$ possible labels.
     $l_1$ is the label with the highest rank and $l_q$ is the label with the lowest rank. 
    \item The labelset size is exponential to the amount of labels: $Y \subseteq \mathbb{P}(L)$.  
    \item Ties in the labelset ranking are allowed, and in some cases all of the labels are tied in first place. A threshold $t \in {1,2,..,q}$ indicates the partitioning of $Y$ into relevant and irrelevant labels. When $t=1$ only the first label in $Y$ is relevant. When $t=q$, or when the threshold is not mentioned, all labels in $Y$ are relevant. When $1 \leq t \leq q$ the labelset is bipartite according to $t$.
    \item The training dataset $\mathcal{D}$ consists of a triplets  $\{ x_i ,Y_i,t_i \}$
    \item The goal is to find a mapping function: $h: \mathcal{X} \to Y$ for a given $t$. 
\end{enumerate}
%explnation (for myself) of the notation: bold-x is an instance which is a vector, has features. Cannot use bold-y since the labels don't have features. using Y_i to indicate that there are multiple labels (but no features to each label)
\end{definition}

In\textbf{ multi-label ranking} problems the labelset is bipartite, i.e., the instance belongs to a ranked subset of the labels, and doesn't belong to the rest of the set. For example, this image contains mainly oranges, apples and bananas in this order, but no pears. 
Several problems are special cases of multi-label ranking:
binary classification, multi-class classification, multi-label classification and label ranking. % citation for label ranking \cite{hullermeier2008label}.
In \textbf{binary classification}, an instance can belong to one of two possible classes, e.g., this image contains either an apple or an orange. 
In \textbf{multi-class} problems, an instance can belong to one out of multiple possible classes, e.g. this image contains either an apple, an orange or a banana. 
In \textbf{multi-label} problems, an instance can belong to many classes (labels), e.g. this image contains an apple and an orange but not a banana. 
In \textbf{label ranking} problems, an instance belongs to a ranked set of classes (labels), e.g. this image contains oranges, apples and bananas, in this order.

Multi-label ranking tasks, as defined in definition \ref{def:mlr}, refer to any problem whose \textit{target class output} is a ranked list of labels and a threshold. Algorithms for solving these tasks are often divided into two sub-groups according to their \textit{target class input}. When the input is a ranked set of labels, it is a label ranking task. When the input is just a set of labels, it is a multi-label task. Formally:

% In this light, the distinction between the sub-cases is characterized as follows:
% A label-ranking (LR) task is a MLR task where there are at least two possible labels, the labelset $Y$ may contain more than one label and is ordered without ties, the threshold is $t=q$ (does not exist). 
% A multi-label (ML) task is a MLR task where there are at least two possible labels, the labelset $Y$ may contain more than one label and may contain ties. The threshold is $1 \leq t \leq q$. 

\begin{definition}\label{def:ML}
\textbf{A multi-label (ML)} task is an MLR  with $1 \leq t \leq q$. The labelset is bipartite according to $t$, with labels $1 \leq t$ tied in first place and considered as relevant.
\end{definition}

\begin{definition}\label{def:LR}
\textbf{A label ranking (LR)} task is an MLR with $t=q$ and no ties in Y. 
\end{definition}

Table \ref{tab:problems} summarizes the differences between MLR and it's sub-tasks according to a few parameters:
\begin{itemize}
    \item Is $m>2$? - There may be many labels available for each instance, or just two.
    \item Is $q>1$? - It may be possible to assign many labels for each instance, or just one. 
    \item Are there ties in $Y$? - Is the labelset completely ordered, or can ties between labels exist?
    \item Does $t=q$? - Are all labels in $Y$ relevant, or does the threshold $t$ partitions the labels into relevant and irrelevant labels?
\end{itemize}
%Herein, we refer to multi-label tasks as any problem where the labelset $Y$ may have multiple labels, regardless of whether these labels are required to be ordered (as in ML), ordered and bipartite (as in MLR) or bipartite.

\begin{table}[h]
\centering
\caption{Table of sub-cases of multi-label ranking}
\label{tab:problems}
\begin{tabular}{@{}lccccc@{}}
\toprule &
\multicolumn{1}{l}{\begin{tabular}[c]{@{}l@{}} $m>2$ \end{tabular}} & 
\multicolumn{1}{l}{\begin{tabular}[c]{@{}l@{}} $q>1$ \end{tabular}} & 
\multicolumn{1}{l}{\begin{tabular}[c]{@{}l@{}} ties in $Y$ \end{tabular}} & 
\multicolumn{1}{l}{\begin{tabular}[c]{@{}l@{}} $t=q$  \end{tabular}}
\\ 
\midrule
Multi-label Ranking (MLR) & yes & yes & yes & yes \\
Label ranking (LR) & yes & yes & yes & no \\
Multi-label (ML) & yes & yes & no & yes \\
Multi-class (MC) & yes & no & no & yes \\
Binary classification & no & no & no & no \\ \bottomrule
\end{tabular}
\end{table}

% Other distinctions between the problems: 
% the instances' features may be either attributes, documents or images. 
% The target input (for learning) may be ordered, multi-label, graph, or hierarchy. 
% \todo{add to the table the required input: order, pairs, partial rankings, graphs}

%Each instance has attributes that can be features, words (in documents), pixels (in images). 

A recent survey categorizes multi-label mining as a sub-problem in multi-target learning \cite{waegeman2019multi}. Related problems in the multi-target domain, that are not covered in this survey, include:
\begin{itemize}
    \item \textbf{Multi variate regression} \cite{borchani2015} and \textbf{dyadic prediction} where the goal is to predict a score for the fit between the instance and the label.  
    \item \textbf{Hierarchical multi-label} \cite{cerri2011hierarchical} where there is explicit side information about the dependencies between the labels. 
    \item\textbf{ Multi-instance} \cite{herrera2016multiple} and \textbf{multi-instance multi-label} \cite{zhou2012multi, huang2018fast} where the training data is composed of a bag of instances that are all assigned the same label or labels. 
    \item \textbf{Multi-view} \cite{zhao2017multi} is similar to multi-instance but the instances may have different feature spaces.% For example, two views can describe a single web page: the text on the web page itself and the anchor text of any web page linking to this web page.
\end{itemize}

\section{Multi-label algorithms}\label{sec:ml}
To date, surveys classify multi-label algorithms as either problem transformation techniques or algorithm adaptation techniques \cite{charte2019snapshot, gibaja2015tutorial, zhang2013review}. 
In the first, the problem is transformed into a simpler single label classification task. In the second, an algorithm used for single label classification is adapted to perform multi-label tasks. Sometimes ensembles techniques are added as a third class of problems and other times they are listed by their underlying base classifier's category (transformation or adaptation). 

However, in the last decade algorithms that are specially designed for multi-label tasks have emerged. 
These algorithms either: try to maximize a specific evaluation measure (e.g. \cite{petterson2011submodular, pillai2017designing, wu2017progressive}), 
focus on a certain sub-task (e.g., feature selection \cite{pereira2018categorizing, spolaor2016systematic, zhang2019distinguishing}),
or attempt to address specific multi-label challenges such as high dimensionality output space, label correlation (e.g.\cite{gu2011correlated, huang2012multi}), imbalance or zero-shot (e.g. \cite{ rios2018few}).
% \cite{gu2011correlated} "There are two challenges in multi-label learning: (1) the labels are interdependent and correlated, and (2) the data are of high dimensionality. In this paper, we aim to tackle these challenges in one shot. In particular, we propose to learn the label correlation and do feature selection simultaneously. We introduce a matrix-variate Normal prior distribution on the weight vectors of the classifier to model the label correlation. Our goal is to find a subset of features, based on which the label correlation regularized loss of label ranking is minimized. "

% The F-score is an harmonic mean of precision and recall \cite{van1974foundation}. 
% \cite{petterson2011submodular} presented  an algorithm which attempts at directly optimising the F-score. It explicitly models the dependencies between pairs of labels in a submodular fashion. Their algorithm for constraint generation which is partially optimal in the sense that all labels it predicts are included in some optimal solution.

Moreover, deep learning algorithms designed specifically for multi-label tasks have been rapidly developing, exhibiting promising results. Indeed, a minority of these algorithms can be classified as algorithm transformation techniques (see e.g. the method suggested in \cite{yang2016exploit}), but more often, the algorithm is not adapted but rather specially tailored for the problem at hand. 

As transformation and adaptation techniques have been exquisitely covered \cite{barot2014review, charte2019snapshot, gibaja2015tutorial, herrera2016multilabel, zhang2018binary, zhang2013review} and as there is only a minor increase in research on them in comparison with other aspects of MLR, 
we lightly scan these foundations (sections \ref{subsec:trans and adapt} and \ref{subsec: ML ensembles}), and then direct our focus on methods from the last demi-decade, specifically on deep learning (section \ref{subsec: DL}) and extreme multi-label methods (section \ref{subsec:XML}). 

%We note that there is also lively and stimulating new research on feature selection

\subsection{Problem adaptation and problem transformation}\label{subsec:trans and adapt}
Problem adaptation techniques were originally suggested for text categorization. However the adaptations soon expanded beyond the text domain and into other scenarios as well. Some of the most noticed adaptations include : Expectation maximization (EM) \cite{mccallum1999multi}, 
SVM \cite{elisseeff2002kernel, joachims1998text, xu2013multi, xu2013fast}, k-NN \cite{calvo2015improving, chiang2012ranking, huang2017multi, skryjomski2019speeding, zhang2007ml}, decision trees \cite{al2014lacova, clare2001}, association rules \cite{thabtah2004mmac} and genetic algorithms \cite{gonccalves2018survey}. 

Problem transformation techniques \cite{barot2014review} focus on transforming the problem into simpler sub-tasks. One classifier is created for each label or pair of labels. The classifiers are then trained separately, and their output is combined. The possible transformations are:
\begin{itemize}
    \item \textbf{Binary Relevance.} The transformation into a binary classification task is known as binary relevance (BR) \cite{zhang2018binary}. 
    The first BR solutions \cite{boutell2004learning, godbole2004discriminative}
    did not consider label correlation. 
    However, many correlation-enabling extensions to binary relevance have been
    proposed in the past decade. These correlations are classified into three sub-classes: first order correlation, pairwise correlation, or full correlation \cite{zhang2013review}.
    \item \textbf{Multi-class.} The most known multi-class transformations are the Label Powerset methods that reduce the problem to a multi-class one by treating each individual labelset as an independent class label \cite{boutell2004learning}. In both BR and multi-class transformations, there is a computational complexity problem, as the solutions do not scale well as the number of labels increase. Thus solutions that reduce the number of classifiers were suggested \cite{madjarov2012two, mencia2008efficient}.
    \item \textbf{Pairwise label comparisons.} Calibrated Label Ranking \cite{furnkranz2008multilabel} transform the dataset into pairs of labels and thus train $k(k−1)/2$  binary classifiers. The output of the classifiers are combined into a ranking of the output labels, with the highest ranked labels considered as relevant. A fictional label can be used to automatically create a bi-partition of the labels into relevant and irrelevant ones \cite{brinker2006unified}. 
\end{itemize}

\subsection{Multi-label ensembles}\label{subsec: ML ensembles}
%There is a nice short survey on ensembles in related work in \cite{papanikolaou2017large}
The 2BR method \cite{tsoumakas2009correlation} uses BR twice and employs stacking. It first learns a BR model, and then builds a second, meta-model that takes the output of the first model and includes a explicit coefficient for correlated labels. The PruDent method focuses on unnecessary label dependencies and error-propagation showing improved results over 2BR \cite{alali2015prudent}.

In Classifier chains (CC) \cite{read2011classifier}, the first classifier is trained on the input attributes. The classifier's output is then added as a new input attribute, and a second classifier is trained, and so on. In this way the classifiers are chained, taking into account the possible label dependencies. In ensembles of classifier chains (ECC) \cite{read2011classifier}, a set of CCs with different orders are trained and the outputs are aggregated.

Hierarchy Of Multi-label classifiERs (HOMER) \cite{tsoumakas2008effective} creates a tree of BR methods, were each leaf contains one label. To classify a new instance, HOMER begin at the root classifier and passes the instance to each child only if the parent predicted any of its labels. The union of the predicted labels by the leaves generates output for the given instance.

AdaBoost.MH \cite{schapire2000boostexter} is the multi-label variation of the well known AdaBoost algorithm \cite{freund1997decision}.   
AdaBoost.MH weighs the labels as well as the instances. Training instances and their corresponding labels that are hard to predict, get incrementally higher weights in following classifiers while instances and labels that are easy to classify get lower weights. This algorithm is designed to minimize the hamming loss. ADTBoost.MH \cite{de2003learning} which uses ADTTrees is an extension of AdaBoost.MH.

Random k-Labelsets (RAkEL) \cite{Tsoumakas2007} selects a number of random k-labelsets and learns a Label Powerset classifer for each of them. These are then aggregated. Enhancements over RAkEL include RAkEL++ \cite{rokach2014ensemble} and RAkELLd \cite{tsoumakas2010random}.

An experimental study on most of the above multi-label ensembles suggests that ECC, followed by RAkEL, exhibit the best overall performance for all of the examined metrics \cite{moyano2018}.

A few notable ensemble methods were published in the last three years. 
ML-FOREST \cite{wu2016ml} builds on Random Forest and  ML-TSVM \cite{chen2016mltsvm} on SVM. 
PRAkEL, a cost-sensitive extension of RAkEL that considers the evaluation criteria and is sensitive to the cost of  misclassifying an instance \cite{wu2017progressive}. 
fRAkEL  speeds up RAkEL by shrinking the samples with irrelevant labels \cite{xu2016multi}.
The TSEN ensemble \cite{zhang2018_TSEN} is based on three-way-decisions \cite{yao2011superiority}. 
The MULE ensemble \cite{papanikolaou2017large} relies on a heterogeneous ensemble that is composed of different base models. The assumption is that different labels can be approximated better by different types of models. MULE incorporates a statistical test to combine the base models. Most of these methods compare their performance to earlier methods, mainly to variations of 2BR, ECC, AdaBoost.MH and RAkEL. However an evaluation of these methods in comparison to one another is still missing.

\subsection{Deep learning methods}\label{subsec: DL}
%summary of sahiner 
Deep learning uses multiple layers to represent the abstractions of data and to 
automatically discover useful features \cite{lecun2015deep, schmidhuber2015deep}. 
Deep learning can cope with large amounts of input features, eliminating the need for feature selection methods.
The learnt feature representations are often accurate for other unseen data as well. While this quality, known as transfer learning \cite{pan2009survey} is not unique to deep learning, in the case of multiple labels it can provide a solution to the label and labelset imbalance problems.
This also means that parameters for a new model (such as number of layers and number of nodes) can be learnt from previous successful models instead of by trial and error. 
Deep learning models are devised using different architectures \cite{pouyanfar2019survey}. 
Convolutional Neural Network (CNN) models \cite{collobert2011natural, lecun2010convolutional} are often used for image processing. 
Recurrent neural network (RNN) models \cite{lai2015recurrent} and their variant Long Short-Term Memory (LSTM) \cite{elman1990finding,tai2015improved} are often used with text and speech. 
Generative adversarial networks (GANs) \cite{goodfellow2014generative} and Restricted Boltzmann Machine (RBM) \cite{ackley1985learning, taylor2007modeling} have been used for unsupervised multi-label learning tasks
%(e.g. detection of food quality using RBM \cite{langkvist2013fast}) 
but we herein focus on a supervised MLR setting as defined in definition \ref{def:mlr}.

Deep learning for multi-label tasks is a growing field, with new papers appearing frequently. We survey some of the recent developments in two main multi-label tasks, which belong to the media domain: image annotation and text annotation. 

\subsubsection{Image annotation}
The growing interest in deep learning for multi-label image annotation is partly driven by new publicly accessible large-scale datasets with quality labels. A few of the most notable ones incluse:
\begin{itemize}
    \item \textbf{The Visual Genome dataset \cite{krishna2017visual}.} Contains over 108K images with an average of 35 labelled objects, 26  attributes, and 21 pairwise relationships between objects. 
    \item \textbf{The ChestX-ray14 dataset \cite{wang2017chestx}.} Contains over 112k chest X-rays from over 30k patients, labeled with up to 14 pathologies or ``No Finding''. %Chest X-rays are a simple and economical medical aid, commonly used for early screening of diseases. 
    \item\textbf{ The MS COCO dataset \cite{lin2014microsoft}.} Contains 328k images with 2.5 million labelled objects, out of a set of 91 labels. 
    \item \textbf{The NUS-WIDE dataset \cite{chua2009nus}.} Contains almost 270k Flickr images and their associated labels, with a total of 5,018 unique labels.
\end{itemize}

Recent reviews of deep learning for medical images highlight the vast amount of emerging research in this field \cite{feng2019deep, karako2018medical, sahiner2019deep}. 
On the chest X-rays dataset, various CNN based methods have been suggested. 
One approach is to transform the problem into multiple single-label classification problems, to which a CNN  architecture is applied \cite{gong2013deep}.
Another suggestion is Hypotheses-CNNPooling (HCP), where a number of object (i.e., image) segment hypotheses are taken as the inputs, then a shared CNN is connected with each hypothesis, and finally the
CNN output results from different hypotheses are aggregated with max pooling to
produce the multi-label predictions \cite{wei2015hcp}. 
The CNN-RNN model uses an underlying RNN model \cite{mikolov2010recurrent} to capture the high order label dependencies. Then, CNN and RNN are combined into one framework to exploit the label dependencies at the global level \cite{wang2016cnn}. 
Other models exist (e.g. \cite{ dong2017learning, guan2018, guan2018diagnose, guendel2018learning,  rajpurkar2017chexnet,shin2016learning}), as well as a cascade ensemble \cite{kumar2018cascade}.
Though multi-instance methods are out of our scope, we note that a multi-instance method, that integrates the images with other information about the patients has recently been reported to enhance performance \cite{baltruschat2019comparison}. 

%should I cut this out?
Various studies have been conducted on non-medical images as well. For a recent review see Voulodimos et al. \cite{voulodimos2018deep}. 
Some of these models focus on learning the label correlations.
A model that learns image-dependent conditional structures \cite{li2016conditional} has been proposed. 
A Spatial Regularization Network (SRN) that captures the spatial correlation between labels as well as the semantic correlation has been suggested \cite{zhu2017learning}.
A feature attention network (FAN) focuses on more important features and learn the correlations among convolutional features \cite{yan2019}.
The Regional Latent Semantic Dependencies (RLSD) model \cite{zhang2018multilabel} specializes in predicting small objects (alongside prediction of large objects) by  first extracting convolutional features, which are further sent to an RPN-like (Regional Proposal Network) localization layer. The layer is designed to localize the regions in an image that may contain multiple semantically dependent labels. These regions are encoded with a fully-connected neural network and further sent to an RNN, which captures the latent semantic dependencies at the regional level. The RNN unit sequentially outputs a multi-class prediction, based on the outputs of the localization layer and the outputs of previous recurrent neurons. Finally, a max-pooling operation is carried out to fuse all the regional outputs as the final prediction. 
Once again, multi-instance methods can enhance performance \cite{luo2019visual}.

%taken from \cite{zhang2018multilabel} (view online, better), works on label dependencies for images
% To model the label dependencies, several approaches have been proposed. The probabilistic graphical models are usually employed in previous works [10], [11], [31]–[33] to model the image feature-label joint distribution. There are several different graph structures to fulfill this purpose. For example, Chow-Liu tree is used to build a tree based on the labels’ mutual dependency in  [31]. In [12], a semi-supervised graph-based method is proposed to utilize single-instance and multi-instance image features for classification. In [33], joint probability is exploited by directed acyclic graph and chain rules. Conditional random field is used in [11], [32] and matrix completion is used in [34] . A limitation of the graph-based approaches is that the richer the label semantic information is, the more complex the graph can be, which causes high computational complexity and low efficiency. Moreover, all of the above methods only model the label dependencies at the global level.

Comparisons of cutting-edge image annotation methods are still lacking.
%from Rokach paper:
A precedent study compared the performance of ten foreground deep learning multi-label APIs on the Visual Genome dataset \cite{kubany2019semantic}. The APIs were evaluated using various metrics. In addition, a semantic similarity metric was used allowing for words with similar meaning to be classified as correct predictions. For example, ``bicycle'' and ``bike'' were both classified as correct for an image of a pedal driven two-wheeler.
The study shows that different APIs excel under different evaluation metrics. Regretfully, the underlying algorithm of each API is not always publicly available. 

% The reviews do not distinguish between multi-label tasks to other tasks. One reports that hundreds of papers or available on PubMed\footnote{ \url{https://www.ncbi.nlm.nih.gov/pubmed}} and that there is an increase in published papers on the subject every year \cite{sahiner2019deep}. 
% While we found only a total of 43 papers that that contain the key words "deep learning" and "multi-label" or "label-ranking", the trend is the same, with one first paper appearing on 2013 and a steady increase to 17 papers in October 2019. 

%(("deep learning"[MeSH Terms] OR ("deep"[All Fields] AND "learning"[All Fields]) OR "deep learning"[All Fields]) OR (deep[All Fields] AND ("neural networks (computer)"[MeSH Terms] OR ("neural"[All Fields] AND "networks"[All Fields] AND "(computer)"[All Fields]) OR "neural networks (computer)"[All Fields] OR ("neural"[All Fields] AND "network"[All Fields]) OR "neural network"[All Fields])) OR (deep[All Fields] AND convolution[All Fields])) AND ((label[All Fields] AND ranking[All Fields]) OR multi-label[All Fields])

\subsubsection{Text Annotation} 
%??? We note that excellent text categorization algorithms that do not employ deep learning exist as well. yes..we covered them earlier. they are old (adaptation and transformation)
%methods
One of the earliest models to employ deep learning for text classification was BP-MLL \cite{zhang2006multilabel}. It formulated multi-label classification problems as a neural network with multiple output nodes, one for each label. 
It was later suggested \cite{nam2014large} to replace the pairwise ranking loss in the model with a cross-entropy loss instead. However these models do not consider label dependencies. 
A CNN model that has a final hidden layer which considers label co-occurence weights  was suggested next \cite{kurata_2016}. The model was analyzed on a small dataset with 103 labels. 
In the CNN-RNN model \cite{chen2017ensemble}, the RNN is set to deal with label co-occurence.
The C2AE algorithm employs a DNN-based label embedding framework and performs joint feature and label embedding \cite{yeh2017learning}.

The Seq2seq model \cite{yang2018sgm} uses a LSTM to generate labels sequentially, and predicts the next label based on its previously predicted labels.
The $LSTM^2$ model \cite{yan2018lstm} utilizes LSTM twice.
The algorithm first builds a representation of the documents in the training set. This is done with a LSTM network that considers word sequences. For each document in the test set, the algorithm search for the most similar documents in the training set and retrieves their labels. The labels are represented as a semantic tree that is trained with dependency parsing. This tree can capture the correlations between labels. Based on the document representation, another LSTM is utilized to rank the document labels. 

Recently, text categorization has also been employed in the context of X-ray images. 
For example, it has been pointed out that China's chest X-ray reports focus more on characterization than on labelling the possible diseases. A recent study uses LSTM to read the X-ray reports, and output labels of pathologies \cite{yan2019combining}. Another study combines the reports and the instances (a multi-instance model) for the same purpose \cite{wang2018tienet}. 

\subsection{Extreme multi-label classification}\label{subsec:XML}
Extreme multi-label (XML) problems, also known as large-scale multi-label problems, refer to problems with an extremely large set of labels, usually in the millions. 
Due to the huge amount of labels, the zero-shot problem, as well as the label and labelset imbalance problems (see section \ref{sec:intro}), are intensified. Deep learning models work well here since they considers both the relatedness of the representations and the context information.

%classification
Currently, all of these problems focus on text classification or on  traditional recommmender system problems that have been reformulated as XML problems \cite{tang2009large, ueda2003parametric}. 
The instances are of one of the following kinds:
\begin{itemize}
    \item \textbf{Text.} Assigning categories (the labels) to a Wikipedia page out of the million categories (labels) available, recommending bid phrases (the labels) to an advertiser with a given ad landing page  \cite{agrawal2013multi}, assigning tags (the labels) to an image.
    In these cases, the instance, be it a web page or an image, is represented by a bag of words. 
    \item \textbf{Item.} Assigning a number of categories (the labels) to an Amazon product item. The instance is represents by item features. 
    \item \textbf{User.} Recommending YouTube videos (the labels) to a user \cite{weston2013label}. The instance is represented by user features. 
\end{itemize}
We note that these tasks often require a ranked output, and that a ranked input might be available. However, to the best of our knowledge, extreme label ranking which explicitly considers ranked inputs (as opposed to extreme multi-label) has not yet been explored. 

PDSparse \cite{yen2016pd} and DiSMEC \cite{babbar2017dismec} tackle the sparseness (high dimensonality) by training one XML model per label. We thus consider them as XML problem transformation methods. 

MLRF \cite{agrawal2013multi} designs a multi-label random forest. To cope with the high dimensonality problem, they assume label independence during the ensemble construction, but they do consider correlations during prediction. 
The FastXML model \cite{prabhu2014fastxml} provides a node partitioning formulation
that improves MLRF results in settings with millions of labels. 
XML-CNN \cite{liu2017deep} uses a CNN model for the actual classification.

The SwiftXML model \cite{prabhu2018extreme} learns label correlation using a word2vec embedding. This allows the discovery of similar labels as these labels are classified closely in the embedded vector. An example given by the authors is that although the labels of the Wikipedia pages of Einstein and Newton are very different, the SwiftXML label embedding will learn that the two are similar, and thus will be able to consider labels given to Einstein's page, also for Newton's page. Authors report to have outperformed SLEEC \cite{bhatia2015sparse} which also employs label embedding and is considered state-of-the-art.
A recent model \cite{zhang2018deep} tackles both feature and label spaces using  
a non-linear embedding based on a graph structure.

% Currently, these problems focus on text classification (the instances are documents, web pages etc.) or recommender systems (the intances are either users or items). 
% Traditional recommmender system problems can be formulated as XML problems \cite{tang2009large, ueda2003parametric}. 
% For example, assigning category labels to a Wikipedia page out of the million categories (labels) available, or assigning a number of categories to an Amazon product. 
% Recommending YouTube videos is another example. Here, the users are the instances, and the task is to predict a  ranked list of videos (the labels) for each user \cite{weston2013label}.
% As a last example consider recommending bid phrases to an advertiser with a given ad landing page. 
% Each web page is treated as an instance, represented by a bag of words, and each query phrase is a label \cite{agrawal2013multi}.  

\section{Label ranking algorithms}\label{sec:lr}
Numerous label ranking algorithms were suggested in the literature.   
One approach is based on turning the problem into several binary classification problems and then combining them into output rankings \cite{cheng2013labelwise, destercke2015cautious, gurrieri2014alternative, har2003constraint, hullermeier2008label}.
Another common approach is based on modifying existing probabilistic algorithms to directly support label ranking. Some main examples are: 
naive Bayes models \cite{aiguzhinov2010similarity}, 
k-nearest neighbor models \cite{brazdil2003ranking} 
and decision tree models such as 
Label Ranking Trees (LRT) \cite{cheng2009decision} and Entropy Based Ranking Trees (ERT) \cite{de2015distance}. 

A few other stand-alone ideas are available as well. 
RPC (Ranking by Pairwise Comparison) \cite{hullermeier2008label} learns pairwise preferences from which a ranking is derived.
Instance Based Logistic Regression (IBLR) combines instance-based learning and logistic regression \cite{cheng2009decision}. Under this approach, the label statistics of neighboring instances are regarded as features by the logistic regression classifier.
A rule based approach learns a reduction technique and provides a mapping
in the form of logical rules \cite{gurrieri2012label}. 
A recent work \cite{korba2018structured} adapts ideas from structured output prediction. They cast the label ranking problem into the structured prediction framework and propose embeddings dedicated to ranking representation. For each embedding they propose a solution to the pre-image problem. This latter suggestion is a harbinger for a bridge between label ranking and structured image prediction. 

Surveys on label ranking algorithms \cite{vembu2010label, zhou2014taxonomy} capture some of the earlier methods.

\subsection{Label ranking ensembles}\label{sec:lr_ensembles}

To the best of our knowledge, only a few papers thus far have investigated the use of ensembles for label ranking \cite{aledo2017tackling, sa2017label, werbin2019beyond, ZHOU2018}.
The ensembles proposed in these papers differ in:
(1) the base label ranking algorithm used,
(2) the method used to sample the data to train each of the simple models (if all models are trained with the exact same data they will output the exact same results, and then there is no need for an ensemble), and
(3) the aggregation method used to combine the results of the simple models.

As the base label ranking algorithm \cite{aledo2017tackling} used Label Ranking Trees (LRT) \cite{cheng2009decision}, \cite{sa2017label} used Ranking Trees (RT) and Entropy Ranking Trees (ERT) \cite{de2015distance}, \cite{ZHOU2018} developed their own method named Top Label As Class (TLAC) and \cite{werbin2019beyond} used both LRT and Ranking by Pairwise Comparison (RPC)  \cite{hullermeier2008label}.

To select the training data for each simple classifier, \cite{aledo2017tackling} and \cite{werbin2019beyond} used a technique known as Bootstrap aggregation or Bagging \cite{breiman1996bagging}: they created $b$ different bags by selecting a subset of the dataset's instances with replacement.
The other two papers \cite{sa2017label,ZHOU2018} suggested modifications to the well-known Random Forest ensemble model \cite{breiman2001random}.
%In Random Forest ensembles, a subset of the dataset's instances is sampled with replacement and a subset of the $m$ features is also sampled. 
As for the aggregation method used, three studies used a voting rule (either Borda or Modal Ranking). The last study \cite{werbin2019beyond} presents VRS (Voting Rule Selector), a meta-model that automatically learns the best voting rule to be used. There four works are summarized in table \ref{table:ensemble_frameworks}.

% \cite{aledo2017tackling} used what they termed as ``majority voting" in order to aggregate the results.
% After attempting to replicate their results, we believe that they actually used Modal Ranking (see \cite{CaragiannisPS14} and section \ref{sec:voting_rules}). \cite{sa2017label} stated that they computed the average rank position of each label and ordered the labels according to their average.
% This is equal to combining the rankings using the Borda voting rule \citep{borda}, which computes the sum of the rank positions for each label and orders the labels according to their sum.
% \cite{ZHOU2018} explicitly stated that they use the Borda voting rule to combine the rankings of the classifiers.

\begin{table}[h]
\begin{center}
\begin{tabular}{ |c||c|c|c|c| } 
\hline
& Aledo et al.(2017) & Sa et al. (2017) & Zhou et al. (2018) & Werbin et al. (2019)\\
 \hline
 Base algorithm & LRT & ERT, RT & TLAC & LRT, RPC\\ 
 Data sampling & Bagging & Random Forest & Random Forest & Bagging\\ 
 Aggregation & Modal ranking & Borda & Borda & Voting Rule Selector \\ 
 \hline
\end{tabular}
\caption{Label ranking ensemble frameworks}
\label{table:ensemble_frameworks}
\end{center}
\end{table}

AdaBoost.MR is the label ranking variation of the well known AdaBoost algorithm \cite{freund1997decision}. The algorithm performs pairwise comparisons between labels. Training instances and their corresponding label pairs that are hard to predict, get incrementally higher weights in following classifiers while instances and label pairs that are easy to classify get lower weights. 
The algorithm is designed to minimize the ranking loss. 
Another boosting-based approach suggests learning a linear utility function for each label, from which the ranking is deduced \cite{dekel2004log}. This approach is more general, as it allows the input to be any sort of preference graph over the labels. A ranking function assigns a utility score to each label. Then, relevant pairwise comparisons as deduced from the graph, are performed. Again, the label weights are updated and additional weight is given to each wrongly ranked pair. As these approaches rely on pairwise comparisons, they do not scale well in the case of many labels.

\section {Evaluation}\label{sec:eval}
Evaluating the performance of multi-label algorithms is difficult.
For example, it is may be impossible to decide which mistake of the following two cases is more serious: one instance with three incorrect labels vs. three instances each with one incorrect label. 
Therefore, a number of performance measures focusing on different aspects have been proposed.  
Schapire and Singer \cite{schapire2000boostexter} initially suggested five metrics: Hamming loss, ranking loss, one-error, average precision. 
The popular micro-F1 and macro-F1 were suggested by Tsoumakas et al. \cite{tsoumakas2010random}. Instance-F1 and AUC measures were discussed by Koyejo et al. \cite{koyejo2015consistent}
A summary of the 11 most common measures is provided Wu et al. \cite{wu2017unified}.
For label ranking, the most popular measure is Kendall-tau. The hamming distance is also often used when permutations represent matching of bipartite graphs. 

%From \cite{gonzalez2019distributed}
% The evaluation metrics for multi-label learning differ from those in traditional classification. The most commonly used metrics are [67] Hamming Loss, subset-accuracy, example-based metrics and label-based metrics. Hamming Loss computes the symmetric difference between the predicted set of labels and the true labels. Subset accuracy requires the predicted set of labels to exactly match the real labels. For both, example-based and label-based metrics, the basic measures are accuracy, precision, recall, and f-measure. They only differ in how they are averaged, hence example-based gives the same weight to each instance, micro-averaged label-based gives more weight to labels with more instances and macro-averaged label-based treats all the labels equally reflecting equally less-represented labels

\subsection{Dataset repositories}
The Mulan repository \cite{mulan} contains 27 multi-label datasets on various subjects. MEKA \cite{MEKA} contains datasets in arff format, suitable for Weka.  An R package automates the use of these datasets \cite{charte16repository}. 
The KDIS research group offers a repository of various datasets obtained from different sources.  \footnote{\url{http://www.uco.es/kdis/mllresources/}}

The Extreme Classification Repository stores 15 extreme multi-label text datasets, as well as code for various algorithms. 
\footnote{\url{http://manikvarma.org/downloads/XC/XMLRepository.html$\#$Prabhu14}}. 
LSHTC series challenge also stores extreme multi-label text datasets \cite{partalasKBAPGAA15}.
The SNAP library \cite{snapnets} contains social and information networks, some of which can (and were) utilized for extreme multi-label tasks (e.g. Amazon products). 
Lastly, it is possible to generate simulated data via a multi-label data generator \cite{tomas2014framework}. 

As for datasets with a ranked target class as input, 
five real world label ranking datasets can be found in \cite{LR_data}. 
The 16 semi-syntetic label ranking datasets used in \cite{cheng10icmla,cheng2009decision}
are stored on the webpage of one of the authors. \footnote{\url{https://cs.uni-paderborn.de/?id=63912}}

\subsection{Stratification of multi-label data}
Estimating the accuracy of a model is traditionally done by splitting the data to training, test, and sometimes also validation subsets. Different techniques are available, such as cross validation, holdout and bootstrap \cite{efron1994introduction, refaeilzadeh2009cross}. 
Random sampling works well when the labels have enough representation in the data. When this is not the case, a stratification approach assures that all subsets contain contain approximately the same proportions of labels as the original dataset. For single-label classification tasks, stratification has been shown to outperform cross-validation with random sampling \cite{ kohavi1995study}. 

For multi-label data, stratification is even more important, as some datasets may have very few patterns representing them and random sampling might place them all in either the training or the test data partitions.

Stratification for multi-label data can either consider the distinct labelsets available in the data, or consider each label separately.
Since the number of labelsets grows exponentially to the number of labels, 
when the number of labels is large there might be only one instance for each example in the labelset, or even no examples at all. For this reason, a method that considers each label separately was suggested by \cite{sechidis2011stratification}. Stratifying the data in this method is a slow procedure, but nonetheless, it improves the classifier performance. 

\section{Research directions and open problems}\label{sec:future}

There are still many paths to explore for multi-label ranking. We outline a few of them here. 

First, in many real-world scenarios, new labels emerge over time. For example, the label ``valentine2030'' will only emerge on year 2030.
Multi-Label learning with emerging new labels has just begun to be considered, and we are aware of only one pioneering work on the subject \cite{zhu2018multi}. 
We have seen herein that recommender system problems can be reformulated as multi-label ones. Perhaps it is time to reconsider the other direction as well \cite{yang2012multilabel}. The ``cold-start'' problem is fundamental in recommender systems. Perhaps their solutions can be adapted to multi-label ranking tasks.
Moreover, evaluation measures need to be fine-tuned to this setting of emerging labels \cite{rios2018few, xian2017zero}. 

Second, a recent study \cite{bogaert2019evaluating} compared recommender system and multi-label classification techniques concluding that AdaBoost with CC chains and BR with multi-label random forest outperform the best recommender system methods in a given cross-selling setting.
However, state-of-the-art multi-label deep learning methods and extreme multi-label methods should be able to do even better but have not been considered in the above study. Moreover, it is interesting to reconsider other recommender and ranking scenarios and reformulate them as MLR tasks as well. 
 
Third, for single-label problems, a set of complexity measures that calculate the overlap and separability of classes has been defined \cite{ho2002complexity}. %, and has lately been enhanced and added to an R package named ECoL (Extended Complexity Library) \cite{lorena2019complex}.
To the best of our knowledge, one characterization metric, called TCS (Theoretical Complexity Score) exists for multi-label tasks \cite{charte2016impact} but no complexity measures exist for label ranking, where the emphasis is on the correct order of labels. 
In a related context, SCUMBLE (Score of ConcUrrence among iMBalanced LabEls) is a new metrics that address label co-occurrence in multi-label datasets \cite{charte2019imbalanced}. It can and should be used to gain better understanding of the available data. 

Fourth, we have reviewed state-of-the-art deep learning algorithms for image annotation and separate algorithms for text classification. One cannot but wonder if these fields can inspire one another and furthermore, if a meta-method for both tasks can be developed.

Lastly, as explained in section \ref{subsec:XML},
extreme label ranking problems where the target input space is ordered have yet to be explored.


\begin{thebibliography}{100}

\bibitem{ackley1985learning}
David~H Ackley, Geoffrey~E Hinton, and Terrence~J Sejnowski.
\newblock A learning algorithm for boltzmann machines.
\newblock {\em Cognitive science}, 9(1):147--169, 1985.

\bibitem{agrawal2013multi}
Rahul Agrawal, Archit Gupta, Yashoteja Prabhu, and Manik Varma.
\newblock Multi-label learning with millions of labels: Recommending advertiser
  bid phrases for web pages.
\newblock In {\em Proceedings of the 22nd international conference on World
  Wide Web}, pages 13--24. ACM, 2013.

\bibitem{aiguzhinov2010similarity}
Artur Aiguzhinov, Carlos Soares, and Ana~Paula Serra.
\newblock A similarity-based adaptation of naive bayes for label ranking:
  Application to the metalearning problem of algorithm recommendation.
\newblock In {\em International Conference on Discovery Science}, pages 16--26.
  Springer, 2010.

\bibitem{al2014lacova}
Reem Al-Otaibi, Meelis Kull, and Peter Flach.
\newblock Lacova: A tree-based multi-label classifier using label covariance as
  splitting criterion.
\newblock In {\em 2014 13th International Conference on Machine Learning and
  Applications}, pages 74--79. IEEE, 2014.

\bibitem{alali2015prudent}
Abdulaziz Alali and Miroslav Kubat.
\newblock Prudent: A pruned and confident stacking approach for multi-label
  classification.
\newblock {\em IEEE Transactions on Knowledge and Data Engineering},
  27(9):2480--2493, 2015.

\bibitem{aledo2017tackling}
Juan~A Aledo, Jos{\'e}~A G{\'a}mez, and David Molina.
\newblock Tackling the supervised label ranking problem by bagging weak
  learners.
\newblock {\em Information Fusion}, 35:38--50, 2017.

\bibitem{babbar2017dismec}
Rohit Babbar and Bernhard Sch{\"o}lkopf.
\newblock Dismec: Distributed sparse machines for extreme multi-label
  classification.
\newblock In {\em Proceedings of the Tenth ACM International Conference on Web
  Search and Data Mining}, pages 721--729. ACM, 2017.

\bibitem{baltruschat2019comparison}
Ivo~M Baltruschat, Hannes Nickisch, Michael Grass, Tobias Knopp, and Axel
  Saalbach.
\newblock Comparison of deep learning approaches for multi-label chest x-ray
  classification.
\newblock {\em Scientific reports}, 9(1):6381, 2019.

\bibitem{barot2014review}
Priyadarshini Barot and Mahesh Panchal.
\newblock Review on various problem transformation methods for classifying
  multi-label data.
\newblock {\em International Journal of Data Mining And Emerging Technologies},
  4(2):45--52, 2014.

\bibitem{bhatia2015sparse}
Kush Bhatia, Himanshu Jain, Purushottam Kar, Manik Varma, and Prateek Jain.
\newblock Sparse local embeddings for extreme multi-label classification.
\newblock In {\em Advances in neural information processing systems}, pages
  730--738, 2015.

\bibitem{bogaert2019evaluating}
Matthias Bogaert, Justine Lootens, Dirk Van~den Poel, and Michel Ballings.
\newblock Evaluating multi-label classifiers and recommender systems in the
  financial service sector.
\newblock {\em European Journal of Operational Research}, 2019.

\bibitem{borchani2015}
Hanen Borchani, Gherardo Varando, Concha Bielza, and Pedro Larrañaga.
\newblock A survey on multi-output regression.
\newblock {\em Wiley Interdisciplinary Reviews: Data Mining and Knowledge
  Discovery}, 5(5):216--233, 2015.

\bibitem{boutell2004learning}
Matthew~R Boutell, Jiebo Luo, Xipeng Shen, and Christopher~M Brown.
\newblock Learning multi-label scene classification.
\newblock {\em Pattern recognition}, 37(9):1757--1771, 2004.

\bibitem{brazdil2003ranking}
Pavel~B Brazdil, Carlos Soares, and Joaquim~Pinto Da~Costa.
\newblock Ranking learning algorithms: Using ibl and meta-learning on accuracy
  and time results.
\newblock {\em Machine Learning}, 50(3):251--277, 2003.

\bibitem{breiman1996bagging}
Leo Breiman.
\newblock Bagging predictors.
\newblock {\em Machine learning}, 24(2):123--140, 1996.

\bibitem{breiman2001random}
Leo Breiman.
\newblock Random forests.
\newblock {\em Machine learning}, 45(1):5--32, 2001.

\bibitem{brinker2006unified}
Klaus Brinker, Johannes F{\"u}rnkranz, and Eyke H{\"u}llermeier.
\newblock A unified model for multilabel classification and ranking.
\newblock In {\em Proceedings of the 2006 conference on ECAI 2006: 17th
  European Conference on Artificial Intelligence August 29--September 1, 2006,
  Riva del Garda, Italy}, pages 489--493. IOS Press, 2006.

\bibitem{bucak2009efficient}
Serhat~S Bucak, Pavan~Kumar Mallapragada, Rong Jin, and Anil~K Jain.
\newblock Efficient multi-label ranking for multi-class learning: application
  to object recognition.
\newblock In {\em 2009 IEEE 12th International Conference on Computer Vision},
  pages 2098--2105. IEEE, 2009.

\bibitem{calvo2015improving}
Jorge Calvo-Zaragoza, Jose~J Valero-Mas, and Juan~R Rico-Juan.
\newblock Improving knn multi-label classification in prototype selection
  scenarios using class proposals.
\newblock {\em Pattern Recognition}, 48(5):1608--1622, 2015.

\bibitem{cerri2011hierarchical}
Ricardo Cerri, Rodrigo~C Barros, and Andr{\'e}~CPLF de~Carvalho.
\newblock Hierarchical multi-label classification for protein function
  prediction: A local approach based on neural networks.
\newblock In {\em 2011 11th International Conference on Intelligent Systems
  Design and Applications}, pages 337--343. IEEE, 2011.

\bibitem{charte2019snapshot}
David Charte, Francisco Charte, Salvador Garc{\'i}a, and Francisco Herrera.
\newblock A snapshot on nonstandard supervised learning problems: taxonomy,
  relationships, problem transformations and algorithm adaptations.
\newblock {\em Progress in Artificial Intelligence}, 8(1):1--14, Apr 2019.

\bibitem{charte-charte:2015}
Francisco Charte and David Charte.
\newblock Working with multilabel datasets in {R}: The mldr package.
\newblock {\em The R Journal}, 7(2):149--162, December 2015.

\bibitem{charte16repository}
Francisco Charte, David Charte, Antonio Rivera, Mar{\'i}a~Jos{\'e} del Jesus,
  and Francisco Herrera.
\newblock R ultimate multilabel dataset repository.
\newblock In Francisco Mart{\'i}nez-{\'A}lvarez, Alicia Troncoso, H{\'e}ctor
  Quinti{\'a}n, and Emilio Corchado, editors, {\em Hybrid Artificial
  Intelligent Systems}, pages 487--499, Cham, 2016. Springer International
  Publishing.

\bibitem{charte2016impact}
Francisco Charte, Antonio Rivera, Mar{\'\i}a~Jos{\'e} del Jesus, and Francisco
  Herrera.
\newblock On the impact of dataset complexity and sampling strategy in
  multilabel classifiers performance.
\newblock In {\em International Conference on Hybrid Artificial Intelligence
  Systems}, pages 500--511. Springer, 2016.

\bibitem{charte2019imbalanced}
Francisco Charte, Antonio~J. Rivera, María~J. del Jesus, and Francisco
  Herrera.
\newblock Dealing with difficult minority labels in imbalanced mutilabel data
  sets.
\newblock {\em Neurocomputing}, 326-327:39 -- 53, 2019.

\bibitem{chen2017ensemble}
Guibin Chen, Deheng Ye, Zhenchang Xing, Jieshan Chen, and Erik Cambria.
\newblock Ensemble application of convolutional and recurrent neural networks
  for multi-label text categorization.
\newblock In {\em 2017 International Joint Conference on Neural Networks
  (IJCNN)}, pages 2377--2383. IEEE, 2017.

\bibitem{chen2016mltsvm}
Wei-Jie Chen, Yuan-Hai Shao, Chun-Na Li, and Nai-Yang Deng.
\newblock Mltsvm: a novel twin support vector machine to multi-label learning.
\newblock {\em Pattern Recognition}, 52:61--74, 2016.

\bibitem{cheng10icmla}
Weiwei Cheng, Krzysztof Dembczy{\'n}ski, and Eyke H\"ullermeier.
\newblock Label ranking methods based on the {P}lackett-{L}uce model.
\newblock In Johannes F{\"u}rnkranz and Thorsten Joachims, editors, {\em
  Proceedings of the 27th International Conference on Machine Learning
  (ICML-10)}, pages 215--222, Haifa, Israel, June 2010. Omnipress.

\bibitem{cheng2013labelwise}
Weiwei Cheng, Sascha Henzgen, and Eyke H{\"u}llermeier.
\newblock Labelwise versus pairwise decomposition in label ranking.
\newblock In {\em LWA}, pages 129--136, 2013.

\bibitem{cheng2009decision}
Weiwei Cheng, Jens H{\"u}hn, and Eyke H{\"u}llermeier.
\newblock Decision tree and instance-based learning for label ranking.
\newblock In {\em Proceedings of the 26th Annual International Conference on
  Machine Learning}, pages 161--168. ACM, 2009.

\bibitem{chiang2012ranking}
Tsung-Hsien Chiang, Hung-Yi Lo, and Shou-De Lin.
\newblock A ranking-based knn approach for multi-label classification.
\newblock In {\em Asian Conference on Machine Learning}, pages 81--96, 2012.

\bibitem{chua2009nus}
Tat-Seng Chua, Jinhui Tang, Richang Hong, Haojie Li, Zhiping Luo, and Yantao
  Zheng.
\newblock Nus-wide: a real-world web image database from national university of
  singapore.
\newblock In {\em Proceedings of the ACM international conference on image and
  video retrieval}, page~48. ACM, 2009.

\bibitem{clare2001}
Amanda Clare and Ross~D. King.
\newblock Knowledge discovery in multi-label phenotype data.
\newblock In Luc De~Raedt and Arno Siebes, editors, {\em Principles of Data
  Mining and Knowledge Discovery}, pages 42--53, Berlin, Heidelberg, 2001.
  Springer Berlin Heidelberg.

\bibitem{collobert2011natural}
Ronan Collobert, Jason Weston, L{\'e}on Bottou, Michael Karlen, Koray
  Kavukcuoglu, and Pavel Kuksa.
\newblock Natural language processing (almost) from scratch.
\newblock {\em Journal of machine learning research}, 12(Aug):2493--2537, 2011.

\bibitem{de2003learning}
Francesco De~Comit{\'e}, R{\'e}mi Gilleron, and Marc Tommasi.
\newblock Learning multi-label alternating decision trees from texts and data.
\newblock In {\em International Workshop on Machine Learning and Data Mining in
  Pattern Recognition}, pages 35--49. Springer, 2003.

\bibitem{de2015distance}
Cl{\'a}udio~Rebelo de~S{\'a}, Carla Rebelo, Carlos Soares, and Arno Knobbe.
\newblock Distance-based decision tree algorithms for label ranking.
\newblock In {\em Portuguese Conference on Artificial Intelligence}, pages
  525--534. Springer, 2015.

\bibitem{dekel2004log}
Ofer Dekel, Yoram Singer, and Christopher~D Manning.
\newblock Log-linear models for label ranking.
\newblock In {\em Advances in neural information processing systems}, pages
  497--504, 2004.

\bibitem{destercke2015cautious}
S{\'e}bastien Destercke, Marie-H{\'e}l{\`e}ne Masson, and Michael Poss.
\newblock Cautious label ranking with label-wise decomposition.
\newblock {\em European Journal of Operational Research}, 246(3):927--935,
  2015.

\bibitem{diplaris2005protein}
Sotiris Diplaris, Grigorios Tsoumakas, Pericles~A Mitkas, and Ioannis Vlahavas.
\newblock Protein classification with multiple algorithms.
\newblock In {\em Panhellenic Conference on Informatics}, pages 448--456.
  Springer, 2005.

\bibitem{dong2017learning}
Yuxi Dong, Yuchao Pan, Jun Zhang, and Wei Xu.
\newblock Learning to read chest x-ray images from 16000+ examples using cnn.
\newblock In {\em Proceedings of the Second IEEE/ACM International Conference
  on Connected Health: Applications, Systems and Engineering Technologies},
  pages 51--57. IEEE Press, 2017.

\bibitem{efron1994introduction}
Bradley Efron and Robert~J Tibshirani.
\newblock {\em An introduction to the bootstrap}.
\newblock CRC press, 1994.

\bibitem{elisseeff2002kernel}
Andr{\'e} Elisseeff and Jason Weston.
\newblock A kernel method for multi-labelled classification.
\newblock In {\em Advances in neural information processing systems}, pages
  681--687, 2002.

\bibitem{elman1990finding}
Jeffrey~L Elman.
\newblock Finding structure in time.
\newblock {\em Cognitive science}, 14(2):179--211, 1990.

\bibitem{feng2019deep}
Yeli Feng, Hui~Seong Teh, and Yiyu Cai.
\newblock Deep learning for chest radiology: A review.
\newblock {\em Current Radiology Reports}, 7(8):24, 2019.

\bibitem{freund1997decision}
Yoav Freund and Robert~E Schapire.
\newblock A decision-theoretic generalization of on-line learning and an
  application to boosting.
\newblock {\em Journal of computer and system sciences}, 55(1):119--139, 1997.

\bibitem{furnkranz2008multilabel}
Johannes F{\"u}rnkranz, Eyke H{\"u}llermeier, Eneldo~Loza Menc{\'\i}a, and
  Klaus Brinker.
\newblock Multilabel classification via calibrated label ranking.
\newblock {\em Machine learning}, 73(2):133--153, 2008.

\bibitem{gibaja2014multi}
Eva Gibaja and Sebasti{\'a}n Ventura.
\newblock Multi-label learning: a review of the state of the art and ongoing
  research.
\newblock {\em Wiley Interdisciplinary Reviews: Data Mining and Knowledge
  Discovery}, 4(6):411--444, 2014.

\bibitem{gibaja2015tutorial}
Eva Gibaja and Sebasti{\'a}n Ventura.
\newblock A tutorial on multilabel learning.
\newblock {\em ACM Computing Surveys (CSUR)}, 47(3):52, 2015.

\bibitem{godbole2004discriminative}
Shantanu Godbole and Sunita Sarawagi.
\newblock Discriminative methods for multi-labeled classification.
\newblock In {\em Pacific-Asia conference on knowledge discovery and data
  mining}, pages 22--30. Springer, 2004.

\bibitem{gonccalves2018survey}
Eduardo~Corr{\^e}a Gon{\c{c}}alves, Alex~A Freitas, and Alexandre Plastino.
\newblock A survey of genetic algorithms for multi-label classification.
\newblock In {\em 2018 IEEE Congress on Evolutionary Computation (CEC)}, pages
  1--8. IEEE, 2018.

\bibitem{gong2013deep}
Yunchao Gong, Yangqing Jia, Thomas Leung, Alexander Toshev, and Sergey Ioffe.
\newblock Deep convolutional ranking for multilabel image annotation.
\newblock In {\em nternational Conference on Learning Representations (ICLR)},
  2014.

\bibitem{goodfellow2014generative}
Ian Goodfellow, Jean Pouget-Abadie, Mehdi Mirza, Bing Xu, David Warde-Farley,
  Sherjil Ozair, Aaron Courville, and Yoshua Bengio.
\newblock Generative adversarial nets.
\newblock In {\em Advances in neural information processing systems}, pages
  2672--2680, 2014.

\bibitem{gu2011correlated}
Quanquan Gu, Zhenhui Li, and Jiawei Han.
\newblock Correlated multi-label feature selection.
\newblock In {\em Proceedings of the 20th ACM international conference on
  Information and knowledge management}, pages 1087--1096. ACM, 2011.

\bibitem{gu2014circlize}
Zuguang Gu, Lei Gu, Roland Eils, Matthias Schlesner, and Benedikt Brors.
\newblock circlize implements and enhances circular visualization in r.
\newblock {\em Bioinformatics}, 30(19):2811--2812, 2014.

\bibitem{guan2018}
Qingji Guan and Yaping Huang.
\newblock Multi-label chest x-ray image classification via category-wise
  residual attention learning.
\newblock {\em Pattern Recognition Letters}, 2018.

\bibitem{guan2018diagnose}
Qingji Guan, Yaping Huang, Zhun Zhong, Zhedong Zheng, Liang Zheng, and Yi~Yang.
\newblock Diagnose like a radiologist: Attention guided convolutional neural
  network for thorax disease classification.
\newblock {\em arXiv preprint arXiv:1801.09927}, 2018.

\bibitem{guendel2018learning}
Sebastian Guendel, Sasa Grbic, Bogdan Georgescu, Siqi Liu, Andreas Maier, and
  Dorin Comaniciu.
\newblock Learning to recognize abnormalities in chest x-rays with
  location-aware dense networks.
\newblock In {\em Iberoamerican Congress on Pattern Recognition}, pages
  757--765. Springer, 2018.

\bibitem{gurrieri2014alternative}
Massimo Gurrieri, Philippe Fortemps, and Xavier Siebert.
\newblock Alternative decomposition techniques for label ranking.
\newblock In {\em International Conference on Information Processing and
  Management of Uncertainty in Knowledge-Based Systems}, pages 464--474.
  Springer, 2014.

\bibitem{gurrieri2012label}
Massimo Gurrieri, Xavier Siebert, Philippe Fortemps, Salvatore Greco, and Roman
  S{\l}owi{\'n}ski.
\newblock Label ranking: A new rule-based label ranking method.
\newblock In {\em International Conference on Information Processing and
  Management of Uncertainty in Knowledge-Based Systems}, pages 613--623.
  Springer, 2012.

\bibitem{har2003constraint}
Sariel Har-Peled, Dan Roth, and Dav Zimak.
\newblock Constraint classification for multiclass classification and ranking.
\newblock In {\em Advances in neural information processing systems}, pages
  809--816, 2003.

\bibitem{herrera2016multilabel}
Francisco Herrera, Francisco Charte, Antonio~J Rivera, and Mar{\'\i}a~J
  Del~Jesus.
\newblock Multilabel classification.
\newblock In {\em Multilabel Classification}, pages 17--31. Springer, 2016.

\bibitem{herrera2016multiple}
Francisco Herrera, Sebasti{\'a}n Ventura, Rafael Bello, Chris Cornelis, Amelia
  Zafra, D{\'a}nel S{\'a}nchez-Tarrag{\'o}, and Sarah Vluymans.
\newblock Multiple instance learning.
\newblock In {\em Multiple instance learning}, pages 17--33. Springer, 2016.

\bibitem{ho2002complexity}
Tin~Kam Ho and Mitra Basu.
\newblock Complexity measures of supervised classification problems.
\newblock {\em IEEE Transactions on Pattern Analysis \& Machine Intelligence},
  24(3):289--300, 2002.

\bibitem{huang2017multi}
Jun Huang, Guorong Li, Shuhui Wang, Zhe Xue, and Qingming Huang.
\newblock Multi-label classification by exploiting local positive and negative
  pairwise label correlation.
\newblock {\em Neurocomputing}, 257:164--174, 2017.

\bibitem{huang2018fast}
Sheng-Jun Huang, Wei Gao, and Zhi-Hua Zhou.
\newblock Fast multi-instance multi-label learning.
\newblock {\em IEEE transactions on pattern analysis and machine intelligence},
  2018.

\bibitem{huang2012multi}
Sheng-Jun Huang and Zhi-Hua Zhou.
\newblock Multi-label learning by exploiting label correlations locally.
\newblock In {\em Twenty-sixth AAAI conference on artificial intelligence},
  2012.

\bibitem{hullermeier2008label}
Eyke H{\"u}llermeier, Johannes F{\"u}rnkranz, Weiwei Cheng, and Klaus Brinker.
\newblock Label ranking by learning pairwise preferences.
\newblock {\em Artificial Intelligence}, 172(16-17):1897--1916, 2008.

\bibitem{ioannou2010}
George Sakkas Grigorios~Tsoumakas Ioannou, Marios and Ioannis Vlahavas.
\newblock Obtaining bipartitions from score vectors for multi-label
  classification.
\newblock In {\em In 2010 22nd IEEE International Conference on Tools with
  Artificial Intelligence}, pages 409--416, 2010.

\bibitem{joachims1998text}
Thorsten Joachims.
\newblock Text categorization with support vector machines: Learning with many
  relevant features.
\newblock In {\em European conference on machine learning}, pages 137--142.
  Springer, 1998.

\bibitem{karako2018medical}
Kenji Karako, Yu~Chen, and Wei Tang.
\newblock On medical application of neural networks trained with various types
  of data.
\newblock {\em Bioscience trends}, 12(6):553--559, 2018.

\bibitem{kohavi1995study}
Ron Kohavi et~al.
\newblock A study of cross-validation and bootstrap for accuracy estimation and
  model selection.
\newblock In {\em Ijcai}, volume~14, pages 1137--1145. Montreal, Canada, 1995.

\bibitem{korba2018structured}
Anna Korba, Alexandre Garcia, and Florence d'Alch{\'e} Buc.
\newblock A structured prediction approach for label ranking.
\newblock In {\em Advances in Neural Information Processing Systems}, pages
  8994--9004, 2018.

\bibitem{koyejo2015consistent}
Oluwasanmi~O Koyejo, Nagarajan Natarajan, Pradeep~K Ravikumar, and Inderjit~S
  Dhillon.
\newblock Consistent multilabel classification.
\newblock In {\em Advances in Neural Information Processing Systems}, pages
  3321--3329, 2015.

\bibitem{krishna2017visual}
Ranjay Krishna, Yuke Zhu, Oliver Groth, Justin Johnson, Kenji Hata, Joshua
  Kravitz, Stephanie Chen, Yannis Kalantidis, Li-Jia Li, David~A Shamma, et~al.
\newblock Visual genome: Connecting language and vision using crowdsourced
  dense image annotations.
\newblock {\em International Journal of Computer Vision}, 123(1):32--73, 2017.

\bibitem{kubany2019semantic}
Adam Kubany, Shimon~Ben Ishay, Ruben-sacha Ohayon, Armin Shmilovici, Lior
  Rokach, and Tomer Doitshman.
\newblock Semantic comparison of state-of-the-art deep learning methods for
  image multi-label classification.
\newblock {\em arXiv preprint arXiv:1903.09190}, 2019.

\bibitem{kumar2018cascade}
Pulkit Kumar, Monika Grewal, and Muktabh~Mayank Srivastava.
\newblock Boosted cascaded convnets for multilabel classification of thoracic
  diseases in chest radiographs.
\newblock In Aur{\'e}lio Campilho, Fakhri Karray, and Bart ter Haar~Romeny,
  editors, {\em Image Analysis and Recognition}, pages 546--552, Cham, 2018.
  Springer International Publishing.

\bibitem{kurata_2016}
Gakuto Kurata, Bing Xiang, and Bowen Zhou.
\newblock Improved neural network-based multi-label classification with better
  initialization leveraging label co-occurrence.
\newblock In {\em Proceedings of the 2016 Conference of the North {A}merican
  Chapter of the Association for Computational Linguistics: Human Language
  Technologies}, pages 521--526, San Diego, California, June 2016. Association
  for Computational Linguistics.

\bibitem{lai2015recurrent}
Siwei Lai, Liheng Xu, Kang Liu, and Jun Zhao.
\newblock Recurrent convolutional neural networks for text classification.
\newblock In {\em Twenty-ninth AAAI conference on artificial intelligence},
  2015.

\bibitem{lecun2015deep}
Yann LeCun, Yoshua Bengio, and Geoffrey Hinton.
\newblock Deep learning.
\newblock {\em nature}, 521(7553):436--444, 2015.

\bibitem{lecun2010convolutional}
Yann LeCun, Koray Kavukcuoglu, and Cl{\'e}ment Farabet.
\newblock Convolutional networks and applications in vision.
\newblock In {\em Proceedings of 2010 IEEE International Symposium on Circuits
  and Systems}, pages 253--256. IEEE, 2010.

\bibitem{snapnets}
Jure Leskovec and Andrej Krevl.
\newblock {SNAP Datasets}: {Stanford} large network dataset collection.
\newblock \url{http://snap.stanford.edu/data}, June 2014.

\bibitem{li2016conditional}
Qiang Li, Maoying Qiao, Wei Bian, and Dacheng Tao.
\newblock Conditional graphical lasso for multi-label image classification.
\newblock In {\em Proceedings of the IEEE Conference on Computer Vision and
  Pattern Recognition}, pages 2977--2986, 2016.

\bibitem{lin2014microsoft}
Tsung-Yi Lin, Michael Maire, Serge Belongie, James Hays, Pietro Perona, Deva
  Ramanan, Piotr Doll{\'a}r, and C~Lawrence Zitnick.
\newblock Microsoft coco: Common objects in context.
\newblock In {\em European conference on computer vision}, pages 740--755.
  Springer, 2014.

\bibitem{liu2017deep}
Jingzhou Liu, Wei-Cheng Chang, Yuexin Wu, and Yiming Yang.
\newblock Deep learning for extreme multi-label text classification.
\newblock In {\em Proceedings of the 40th International ACM SIGIR Conference on
  Research and Development in Information Retrieval}, pages 115--124. ACM,
  2017.

\bibitem{luo2019visual}
Yan Luo, Ming Jiang, and Qi~Zhao.
\newblock Visual attention in multi-label image classification.
\newblock In {\em Proceedings of the IEEE Conference on Computer Vision and
  Pattern Recognition Workshops}, pages 0--0, 2019.

\bibitem{madjarov2012two}
Gjorgji Madjarov, Dejan Gjorgjevikj, and Sa{\v{s}}o D{\v{z}}eroski.
\newblock Two stage architecture for multi-label learning.
\newblock {\em Pattern Recognition}, 45(3):1019--1034, 2012.

\bibitem{mccallum1999multi}
Andrew McCallum.
\newblock Multi-label text classification with a mixture model trained by em.
\newblock In {\em AAAI workshop on Text Learning}, pages 1--7, 1999.

\bibitem{mencia2008efficient}
Eneldo~Loza Mencia and Johannes F{\"u}rnkranz.
\newblock Efficient pairwise multilabel classification for large-scale problems
  in the legal domain.
\newblock In {\em Joint European Conference on Machine Learning and Knowledge
  Discovery in Databases}, pages 50--65. Springer, 2008.

\bibitem{mikolov2010recurrent}
Tom{\'a}{\v{s}} Mikolov, Martin Karafi{\'a}t, Luk{\'a}{\v{s}} Burget, Jan
  {\v{C}}ernock{\`y}, and Sanjeev Khudanpur.
\newblock Recurrent neural network based language model.
\newblock In {\em Eleventh annual conference of the international speech
  communication association}, 2010.

\bibitem{mironczuk2018}
Marcin~Michał Mirończuk and Jarosław Protasiewicz.
\newblock A recent overview of the state-of-the-art elements of text
  classification.
\newblock {\em Expert Systems with Applications}, 106:36 -- 54, 2018.

\bibitem{moyano2018}
Jose~M. Moyano, Eva~L. Gibaja, Krzysztof~J. Cios, and Sebastián Ventura.
\newblock Review of ensembles of multi-label classifiers: Models, experimental
  study and prospects.
\newblock {\em Information Fusion}, 44:33 -- 45, 2018.

\bibitem{mujtaba2019}
Ghulam Mujtaba, Liyana Shuib, Norisma Idris, Wai~Lam Hoo, Ram~Gopal Raj, Kamran
  Khowaja, Khairunisa Shaikh, and Henry~Friday Nweke.
\newblock Clinical text classification research trends: Systematic literature
  review and open issues.
\newblock {\em Expert Systems with Applications}, 116:494 -- 520, 2019.

\bibitem{nam2014large}
Jinseok Nam, Jungi Kim, Eneldo~Loza Menc{\'\i}a, Iryna Gurevych, and Johannes
  F{\"u}rnkranz.
\newblock Large-scale multi-label text classification—revisiting neural
  networks.
\newblock In {\em Joint european conference on machine learning and knowledge
  discovery in databases}, pages 437--452. Springer, 2014.

\bibitem{pan2009survey}
Sinno~Jialin Pan and Qiang Yang.
\newblock A survey on transfer learning.
\newblock {\em IEEE Transactions on knowledge and data engineering},
  22(10):1345--1359, 2009.

\bibitem{papanikolaou2017large}
Yannis Papanikolaou, Grigorios Tsoumakas, Manos Laliotis, Nikos Markantonatos,
  and Ioannis Vlahavas.
\newblock Large-scale online semantic indexing of biomedical articles via an
  ensemble of multi-label classification models.
\newblock {\em Journal of biomedical semantics}, 8(1):43, 2017.

\bibitem{partalasKBAPGAA15}
Ioannis Partalas, Aris Kosmopoulos, Nicolas Baskiotis, Thierry Arti{\`{e}}res,
  George Paliouras, {\'{E}}ric Gaussier, Ion Androutsopoulos, Massih{-}Reza
  Amini, and Patrick Gallinari.
\newblock {LSHTC:} {A} benchmark for large-scale text classification.
\newblock {\em CoRR}, abs/1503.08581, 2015.

\bibitem{pereira2018categorizing}
Rafael~B Pereira, Alexandre Plastino, Bianca Zadrozny, and Luiz~HC Merschmann.
\newblock Categorizing feature selection methods for multi-label
  classification.
\newblock {\em Artificial Intelligence Review}, 49(1):57--78, 2018.

\bibitem{petterson2011submodular}
James Petterson and Tib{\'e}rio~S Caetano.
\newblock Submodular multi-label learning.
\newblock In {\em Advances in Neural Information Processing Systems}, pages
  1512--1520, 2011.

\bibitem{pillai2017designing}
Ignazio Pillai, Giorgio Fumera, and Fabio Roli.
\newblock Designing multi-label classifiers that maximize f measures: State of
  the art.
\newblock {\em Pattern Recognition}, 61:394--404, 2017.

\bibitem{pouyanfar2019survey}
Samira Pouyanfar, Saad Sadiq, Yilin Yan, Haiman Tian, Yudong Tao, Maria~Presa
  Reyes, Mei-Ling Shyu, Shu-Ching Chen, and SS~Iyengar.
\newblock A survey on deep learning: Algorithms, techniques, and applications.
\newblock {\em ACM Computing Surveys (CSUR)}, 51(5):92, 2019.

\bibitem{prabhu2018extreme}
Yashoteja Prabhu, Anil Kag, Shilpa Gopinath, Kunal Dahiya, Shrutendra Harsola,
  Rahul Agrawal, and Manik Varma.
\newblock Extreme multi-label learning with label features for warm-start
  tagging, ranking \& recommendation.
\newblock In {\em Proceedings of the Eleventh ACM International Conference on
  Web Search and Data Mining}, pages 441--449. ACM, 2018.

\bibitem{prabhu2014fastxml}
Yashoteja Prabhu and Manik Varma.
\newblock Fastxml: A fast, accurate and stable tree-classifier for extreme
  multi-label learning.
\newblock In {\em Proceedings of the 20th ACM SIGKDD international conference
  on Knowledge discovery and data mining}, pages 263--272. ACM, 2014.

\bibitem{rajpurkar2017chexnet}
Pranav Rajpurkar, Jeremy Irvin, Kaylie Zhu, Brandon Yang, Hershel Mehta, Tony
  Duan, Daisy Ding, Aarti Bagul, Curtis Langlotz, Katie Shpanskaya, et~al.
\newblock Chexnet: Radiologist-level pneumonia detection on chest x-rays with
  deep learning.
\newblock {\em arXiv preprint arXiv:1711.05225}, 2017.

\bibitem{read2011classifier}
Jesse Read, Bernhard Pfahringer, Geoff Holmes, and Eibe Frank.
\newblock Classifier chains for multi-label classification.
\newblock {\em Machine learning}, 85(3):333, 2011.

\bibitem{MEKA}
Jesse Read, Peter Reutemann, Bernhard Pfahringer, and Geoff Holmes.
\newblock {MEKA}: A multi-label/multi-target extension to {Weka}.
\newblock {\em Journal of Machine Learning Research}, 17(21):1--5, 2016.

\bibitem{LR_data}
Claudio Rebelo~de Sa.
\newblock Label ranking datasets.
\newblock \url{ http://dx.doi.org/10.17632/3mv94c8jpc.2 }, 2018.
\newblock Accessed: 2019-10-30.

\bibitem{refaeilzadeh2009cross}
Payam Refaeilzadeh, Lei Tang, and Huan Liu.
\newblock Cross-validation.
\newblock {\em Encyclopedia of database systems}, pages 532--538, 2009.

\bibitem{rios2018few}
Anthony Rios and Ramakanth Kavuluru.
\newblock Few-shot and zero-shot multi-label learning for structured label
  spaces.
\newblock In {\em Proceedings of the Conference on Empirical Methods in Natural
  Language Processing. Conference on Empirical Methods in Natural Language
  Processing}, volume 2018, page 3132. NIH Public Access, 2018.

\bibitem{rokach2014ensemble}
Lior Rokach, Alon Schclar, and Ehud Itach.
\newblock Ensemble methods for multi-label classification.
\newblock {\em Expert Systems with Applications}, 41(16):7507--7523, 2014.

\bibitem{rubin2012statistical}
Timothy~N Rubin, America Chambers, Padhraic Smyth, and Mark Steyvers.
\newblock Statistical topic models for multi-label document classification.
\newblock {\em Machine learning}, 88(1-2):157--208, 2012.

\bibitem{sa2017label}
Cl{\'a}udio~Rebelo S{\'a}, Carlos Soares, Arno Knobbe, and Paulo Cortez.
\newblock Label ranking forests.
\newblock {\em Expert Systems}, 34(1), 2017.

\bibitem{sahiner2019deep}
Berkman Sahiner, Aria Pezeshk, Lubomir~M Hadjiiski, Xiaosong Wang, Karen
  Drukker, Kenny~H Cha, Ronald~M Summers, and Maryellen~L Giger.
\newblock Deep learning in medical imaging and radiation therapy.
\newblock {\em Medical physics}, 46(1):e1--e36, 2019.

\bibitem{schapire2000boostexter}
Robert~E Schapire and Yoram Singer.
\newblock Boostexter: A boosting-based system for text categorization.
\newblock {\em Machine learning}, 39(2-3):135--168, 2000.

\bibitem{schmidhuber2015deep}
J{\"u}rgen Schmidhuber.
\newblock Deep learning in neural networks: An overview.
\newblock {\em Neural networks}, 61:85--117, 2015.

\bibitem{sechidis2011stratification}
Konstantinos Sechidis, Grigorios Tsoumakas, and Ioannis Vlahavas.
\newblock On the stratification of multi-label data.
\newblock In {\em Joint European Conference on Machine Learning and Knowledge
  Discovery in Databases}, pages 145--158. Springer, 2011.

\bibitem{shin2016learning}
Hoo-Chang Shin, Kirk Roberts, Le~Lu, Dina Demner-Fushman, Jianhua Yao, and
  Ronald~M Summers.
\newblock Learning to read chest x-rays: Recurrent neural cascade model for
  automated image annotation.
\newblock In {\em Proceedings of the IEEE conference on computer vision and
  pattern recognition}, pages 2497--2506, 2016.

\bibitem{skryjomski2019speeding}
Przemys{\l}aw Skryjomski, Bartosz Krawczyk, and Alberto Cano.
\newblock Speeding up k-nearest neighbors classifier for large-scale
  multi-label learning on gpus.
\newblock {\em Neurocomputing}, 354:10--19, 2019.

\bibitem{spolaor2016systematic}
Newton Spola{\^o}r, Maria~Carolina Monard, Grigorios Tsoumakas, and Huei~Diana
  Lee.
\newblock A systematic review of multi-label feature selection and a new method
  based on label construction.
\newblock {\em Neurocomputing}, 180:3--15, 2016.

\bibitem{tai2015improved}
Kai~Sheng Tai, Richard Socher, and Christopher~D Manning.
\newblock Improved semantic representations from tree-structured long
  short-term memory networks.
\newblock In {\em Proceedings of the 53rd Annual Meeting of the Association for
  Computational Linguistics and the 7th International Joint Conference on
  Natural Language Processing (Volume 1: Long Papers)}, pages 1556--1566, 2015.

\bibitem{tang2009large}
Lei Tang, Suju Rajan, and Vijay~K Narayanan.
\newblock Large scale multi-label classification via metalabeler.
\newblock In {\em Proceedings of the 18th international conference on World
  wide web}, pages 211--220. ACM, 2009.

\bibitem{taylor2007modeling}
Graham~W Taylor, Geoffrey~E Hinton, and Sam~T Roweis.
\newblock Modeling human motion using binary latent variables.
\newblock In {\em Advances in neural information processing systems}, pages
  1345--1352, 2007.

\bibitem{thabtah2004mmac}
Fadi~A Thabtah, Peter Cowling, and Yonghong Peng.
\newblock Mmac: A new multi-class, multi-label associative classification
  approach.
\newblock In {\em Fourth IEEE International Conference on Data Mining
  (ICDM'04)}, pages 217--224. IEEE, 2004.

\bibitem{tomas2014framework}
Jimena~Torres Tom{\'a}s, Newton Spola{\^o}r, Everton~Alvares Cherman, and
  Maria~Carolina Monard.
\newblock A framework to generate synthetic multi-label datasets.
\newblock {\em Electronic Notes in Theoretical Computer Science}, 302:155--176,
  2014.

\bibitem{trohidis2008multi}
Konstantinos Trohidis, Grigorios Tsoumakas, George Kalliris, and Ioannis~P
  Vlahavas.
\newblock Multi-label classification of music into emotions.
\newblock In {\em ISMIR}, volume~8, pages 325--330, 2008.

\bibitem{tsoumakas2009correlation}
Grigorios Tsoumakas, Anastasios Dimou, Eleftherios Spyromitros, Vasileios
  Mezaris, Ioannis Kompatsiaris, and Ioannis Vlahavas.
\newblock Correlation-based pruning of stacked binary relevance models for
  multi-label learning.
\newblock In {\em Proceedings of the 1st international workshop on learning
  from multi-label data}, pages 101--116, 2009.

\bibitem{tsoumakas2008effective}
Grigorios Tsoumakas, Ioannis Katakis, and Ioannis Vlahavas.
\newblock Effective and efficient multilabel classification in domains with
  large number of labels.
\newblock In {\em Proc. ECML/PKDD 2008 Workshop on Mining Multidimensional Data
  (MMD’08)}, volume~21, pages 53--59. sn, 2008.

\bibitem{tsoumakas2010random}
Grigorios Tsoumakas, Ioannis Katakis, and Ioannis Vlahavas.
\newblock Random k-labelsets for multilabel classification.
\newblock {\em IEEE Transactions on Knowledge and Data Engineering},
  23(7):1079--1089, 2010.

\bibitem{mulan}
Grigorios Tsoumakas, Eleftherios Spyromitros-Xioufis, Jozef Vilcek, and Ioannis
  Vlahavas.
\newblock Mulan: A java library for multi-label learning.
\newblock {\em Journal of Machine Learning Research}, 12:2411--2414, 2011.

\bibitem{Tsoumakas2007}
Grigorios Tsoumakas and Ioannis Vlahavas.
\newblock Random k-labelsets: An ensemble method for multilabel classification.
\newblock In Joost~N. Kok, Jacek Koronacki, Raomon Lopez~de Mantaras, Stan
  Matwin, Dunja Mladeni{\v{c}}, and Andrzej Skowron, editors, {\em Machine
  Learning: ECML 2007}, pages 406--417, Berlin, Heidelberg, 2007. Springer
  Berlin Heidelberg.

\bibitem{ueda2003parametric}
Naonori Ueda and Kazumi Saito.
\newblock Parametric mixture models for multi-labeled text.
\newblock In {\em Advances in neural information processing systems}, pages
  737--744, 2003.

\bibitem{vembu2010label}
Shankar Vembu and Thomas G{\"a}rtner.
\newblock Label ranking algorithms: A survey.
\newblock In {\em Preference learning}, pages 45--64. Springer, 2010.

\bibitem{voulodimos2018deep}
Athanasios Voulodimos, Nikolaos Doulamis, Anastasios Doulamis, and Eftychios
  Protopapadakis.
\newblock Deep learning for computer vision: A brief review.
\newblock {\em Computational intelligence and neuroscience}, 2018, 2018.

\bibitem{waegeman2019multi}
Willem Waegeman, Krzysztof Dembczy{\'n}ski, and Eyke H{\"u}llermeier.
\newblock Multi-target prediction: a unifying view on problems and methods.
\newblock {\em Data Mining and Knowledge Discovery}, 33(2):293--324, 2019.

\bibitem{wang2016cnn}
Jiang Wang, Yi~Yang, Junhua Mao, Zhiheng Huang, Chang Huang, and Wei Xu.
\newblock Cnn-rnn: A unified framework for multi-label image classification.
\newblock In {\em Proceedings of the IEEE conference on computer vision and
  pattern recognition}, pages 2285--2294, 2016.

\bibitem{wang2013multi}
Xi~Wang and Gita Sukthankar.
\newblock Multi-label relational neighbor classification using social context
  features.
\newblock In {\em Proceedings of the 19th ACM SIGKDD international conference
  on Knowledge discovery and data mining}, pages 464--472. ACM, 2013.

\bibitem{wang2017chestx}
Xiaosong Wang, Yifan Peng, Le~Lu, Zhiyong Lu, Mohammadhadi Bagheri, and
  Ronald~M Summers.
\newblock Chestx-ray8: Hospital-scale chest x-ray database and benchmarks on
  weakly-supervised classification and localization of common thorax diseases.
\newblock In {\em Proceedings of the IEEE conference on computer vision and
  pattern recognition}, pages 2097--2106, 2017.

\bibitem{wang2018tienet}
Xiaosong Wang, Yifan Peng, Le~Lu, Zhiyong Lu, and Ronald~M Summers.
\newblock Tienet: Text-image embedding network for common thorax disease
  classification and reporting in chest x-rays.
\newblock In {\em Proceedings of the IEEE conference on computer vision and
  pattern recognition}, pages 9049--9058, 2018.

\bibitem{wei2015hcp}
Yunchao Wei, Wei Xia, Min Lin, Junshi Huang, Bingbing Ni, Jian Dong, Yao Zhao,
  and Shuicheng Yan.
\newblock Hcp: A flexible cnn framework for multi-label image classification.
\newblock {\em IEEE transactions on pattern analysis and machine intelligence},
  38(9):1901--1907, 2015.

\bibitem{werbin2019beyond}
Havi Werbin-Ofir, Lihi Dery, and Erez Shmueli.
\newblock Beyond majority: Label ranking ensembles based on voting rules.
\newblock {\em Expert Systems with Applications}, 2019.

\bibitem{weston2013label}
Jason Weston, Ameesh Makadia, and Hector Yee.
\newblock Label partitioning for sublinear ranking.
\newblock In {\em International Conference on Machine Learning}, pages
  181--189, 2013.

\bibitem{wu2016ml}
Qingyao Wu, Mingkui Tan, Hengjie Song, Jian Chen, and Michael~K Ng.
\newblock Ml-forest: A multi-label tree ensemble method for multi-label
  classification.
\newblock {\em IEEE transactions on knowledge and data engineering},
  28(10):2665--2680, 2016.

\bibitem{wu2017unified}
Xi-Zhu Wu and Zhi-Hua Zhou.
\newblock A unified view of multi-label performance measures.
\newblock In {\em Proceedings of the 34th International Conference on Machine
  Learning-Volume 70}, pages 3780--3788. JMLR. org, 2017.

\bibitem{wu2017progressive}
Yu-Ping Wu and Hsuan-Tien Lin.
\newblock Progressive random k-labelsets for cost-sensitive multi-label
  classification.
\newblock {\em Machine Learning}, 106(5):671--694, 2017.

\bibitem{xian2017zero}
Yongqin Xian, Bernt Schiele, and Zeynep Akata.
\newblock Zero-shot learning-the good, the bad and the ugly.
\newblock In {\em Proceedings of the IEEE Conference on Computer Vision and
  Pattern Recognition}, pages 4582--4591, 2017.

\bibitem{xu2013fast}
Jianhua Xu.
\newblock Fast multi-label core vector machine.
\newblock {\em Pattern Recognition}, 46(3):885--898, 2013.

\bibitem{xu2016multi}
Suping Xu, Xibei Yang, Hualong Yu, Dong-Jun Yu, Jingyu Yang, and Eric~CC Tsang.
\newblock Multi-label learning with label-specific feature reduction.
\newblock {\em Knowledge-Based Systems}, 104:52--61, 2016.

\bibitem{xu2013multi}
Yan Xu, Liping Jiao, Siyu Wang, Junsheng Wei, Yubo Fan, Maode Lai, and Eric
  I-chao Chang.
\newblock Multi-label classification for colon cancer using histopathological
  images.
\newblock {\em Microscopy Research and Technique}, 76(12):1266--1277, 2013.

\bibitem{yan2019combining}
Fengqi Yan, Xin Huang, Yao Yao, Mingming Lu, and Maozhen Li.
\newblock Combining lstm and densenet for automatic annotation and
  classification of chest x-ray images.
\newblock {\em IEEE Access}, 7:74181--74189, 2019.

\bibitem{yan2018lstm}
Yan Yan, Ying Wang, Wen-Chao Gao, Bo-Wen Zhang, Chun Yang, and Xu-Cheng Yin.
\newblock Lstm: Multi-label ranking for document classification.
\newblock {\em Neural Processing Letters}, 47(1):117--138, 2018.

\bibitem{yan2019}
Z.~{Yan}, W.~{Liu}, S.~{Wen}, and Y.~{Yang}.
\newblock Multi-label image classification by feature attention network.
\newblock {\em IEEE Access}, 7:98005--98013, 2019.

\bibitem{yang2016exploit}
Hao Yang, Joey Tianyi~Zhou, Yu~Zhang, Bin-Bin Gao, Jianxin Wu, and Jianfei Cai.
\newblock Exploit bounding box annotations for multi-label object recognition.
\newblock In {\em Proceedings of the IEEE Conference on Computer Vision and
  Pattern Recognition}, pages 280--288, 2016.

\bibitem{yang2018sgm}
Pengcheng Yang, Xu~Sun, Wei Li, Shuming Ma, Wei Wu, and Houfeng Wang.
\newblock Sgm: Sequence generation model for multi-label classification.
\newblock In {\em Proceedings of the 27th International Conference on
  Computational Linguistics}, pages 3915--3926, 2018.

\bibitem{yang2012multilabel}
Yiming Yang and Siddharth Gopal.
\newblock Multilabel classification with meta-level features in a
  learning-to-rank framework.
\newblock {\em Machine Learning}, 88(1-2):47--68, 2012.

\bibitem{yao2011superiority}
Yiyu Yao.
\newblock The superiority of three-way decisions in probabilistic rough set
  models.
\newblock {\em Information Sciences}, 181(6):1080--1096, 2011.

\bibitem{yeh2017learning}
Chih-Kuan Yeh, Wei-Chieh Wu, Wei-Jen Ko, and Yu-Chiang~Frank Wang.
\newblock Learning deep latent space for multi-label classification.
\newblock In {\em Thirty-First AAAI Conference on Artificial Intelligence},
  2017.

\bibitem{yen2016pd}
Ian En-Hsu Yen, Xiangru Huang, Pradeep Ravikumar, Kai Zhong, and Inderjit
  Dhillon.
\newblock Pd-sparse: A primal and dual sparse approach to extreme multiclass
  and multilabel classification.
\newblock In {\em International Conference on Machine Learning}, pages
  3069--3077, 2016.

\bibitem{zhang2018multilabel}
Junjie Zhang, Qi~Wu, Chunhua Shen, Jian Zhang, and Jianfeng Lu.
\newblock Multilabel image classification with regional latent semantic
  dependencies.
\newblock {\em IEEE Transactions on Multimedia}, 20(10):2801--2813, 2018.

\bibitem{zhang2018binary}
Min-Ling Zhang, Yu-Kun Li, Xu-Ying Liu, and Xin Geng.
\newblock Binary relevance for multi-label learning: an overview.
\newblock {\em Frontiers of Computer Science}, 12(2):191--202, 2018.

\bibitem{zhang2006multilabel}
Min-Ling Zhang and Zhi-Hua Zhou.
\newblock Multilabel neural networks with applications to functional genomics
  and text categorization.
\newblock {\em IEEE transactions on Knowledge and Data Engineering},
  18(10):1338--1351, 2006.

\bibitem{zhang2007ml}
Min-Ling Zhang and Zhi-Hua Zhou.
\newblock Ml-knn: A lazy learning approach to multi-label learning.
\newblock {\em Pattern recognition}, 40(7):2038--2048, 2007.

\bibitem{zhang2013review}
Min-Ling Zhang and Zhi-Hua Zhou.
\newblock A review on multi-label learning algorithms.
\newblock {\em IEEE transactions on knowledge and data engineering},
  26(8):1819--1837, 2013.

\bibitem{zhang2019distinguishing}
Ping Zhang, Guixia Liu, and Wanfu Gao.
\newblock Distinguishing two types of labels for multi-label feature selection.
\newblock {\em Pattern Recognition}, 2019.

\bibitem{zhang2018deep}
Wenjie Zhang, Junchi Yan, Xiangfeng Wang, and Hongyuan Zha.
\newblock Deep extreme multi-label learning.
\newblock In {\em Proceedings of the 2018 ACM on International Conference on
  Multimedia Retrieval}, pages 100--107. ACM, 2018.

\bibitem{zhang2018_TSEN}
Yuanjian Zhang, Duoqian Miao, Zhifei Zhang, Jianfeng Xu, and Sheng Luo.
\newblock A three-way selective ensemble model for multi-label classification.
\newblock {\em International Journal of Approximate Reasoning}, 103:394 -- 413,
  2018.

\bibitem{zhao2017multi}
Jing Zhao, Xijiong Xie, Xin Xu, and Shiliang Sun.
\newblock Multi-view learning overview: Recent progress and new challenges.
\newblock {\em Information Fusion}, 38:43--54, 2017.

\bibitem{zhou2014taxonomy}
Yangming Zhou, Yangguang Liu, Jiangang Yang, Xiaoqi He, and Liangliang Liu.
\newblock A taxonomy of label ranking algorithms.
\newblock {\em Journal of Computers}, 9(3):557--565, 2014.

\bibitem{ZHOU2018}
Yangming Zhou and Guoping Qiu.
\newblock Random forest for label ranking.
\newblock {\em Expert Systems with Applications}, 112:99 -- 109, 2018.

\bibitem{zhou2012multi}
Zhi-Hua Zhou, Min-Ling Zhang, Sheng-Jun Huang, and Yu-Feng Li.
\newblock Multi-instance multi-label learning.
\newblock {\em Artificial Intelligence}, 176(1):2291--2320, 2012.

\bibitem{zhu2017learning}
Feng Zhu, Hongsheng Li, Wanli Ouyang, Nenghai Yu, and Xiaogang Wang.
\newblock Learning spatial regularization with image-level supervisions for
  multi-label image classification.
\newblock In {\em Proceedings of the IEEE Conference on Computer Vision and
  Pattern Recognition}, pages 5513--5522, 2017.

\bibitem{zhu2018multi}
Yue Zhu, Kai~Ming Ting, and Zhi-Hua Zhou.
\newblock Multi-label learning with emerging new labels.
\newblock {\em IEEE Transactions on Knowledge and Data Engineering},
  30(10):1901--1914, 2018.

\end{thebibliography}
\end{document}